\documentclass[nohyperref]{article}

\usepackage{microtype}
\usepackage{graphicx}
\usepackage{subfigure}
\usepackage{booktabs} 

\usepackage{hyperref}
\usepackage{url}
\usepackage[accepted]{icml2022}

\usepackage{amsmath}
\usepackage{amssymb}
\usepackage{mathtools}
\usepackage{amsthm}
\usepackage{amsfonts, dsfont}

\usepackage[capitalize,noabbrev]{cleveref}

\theoremstyle{plain}
\newtheorem{theorem}{Theorem}[section]

\newtheorem{corollary}[theorem]{Corollary}
\theoremstyle{definition}

\theoremstyle{remark}

\usepackage[textsize=tiny]{todonotes}
\usepackage{nicefrac}       
\usepackage{multirow}
\usepackage{xcolor}         
\usepackage{color}
\usepackage{soul}
\usepackage{bm}
\usepackage{tabularx}
\usepackage{graphicx}
\usepackage{wrapfig}
\usepackage{subfigure}

\usepackage{algpseudocode}

\crefname{ineq}{inequality}{inequalities}
\newcommand{\norm}[1]{\left\lVert#1\right\rVert}

\newcommand{\inner}[1]{\langle#1\rangle}
\newcommand{\red}[1]{\textcolor{black}{#1}}

\icmltitlerunning{Self-Supervised Representation Learning via Latent Graph Prediction}

\begin{document}

\twocolumn[
\icmltitle{Self-Supervised Representation Learning via Latent Graph Prediction}



\icmlsetsymbol{equal}{*}

\begin{icmlauthorlist}
\icmlauthor{Yaochen Xie}{equal,tamu}
\icmlauthor{Zhao Xu}{equal,tamu}
\icmlauthor{Shuiwang Ji}{tamu}
\end{icmlauthorlist}

\icmlaffiliation{tamu}{Department of Computer Science \& Engineering, Texas A\&M University, College Station, USA}

\icmlcorrespondingauthor{Yaochen Xie}{ethanycx@tamu.edu}
\icmlcorrespondingauthor{Shuiwang Ji}{sji@tamu.edu}

\icmlkeywords{Self-supervised learning, graph neural networks, deep learning}

\vskip 0.3in
]



\printAffiliationsAndNotice{\icmlEqualContribution} 

\begin{abstract}
Self-supervised learning (SSL) of graph neural networks is emerging as a promising way of leveraging unlabeled data. Currently, most methods are based on contrastive learning adapted from the image domain, which requires view generation and a sufficient number of negative samples. In contrast, existing predictive models do not require negative sampling, but lack theoretical guidance on the design of pretext training tasks. In this work, we propose the \textit{LaGraph}, a theoretically grounded predictive SSL framework based on latent graph prediction. Learning objectives of \textit{LaGraph} are derived as self-supervised upper bounds to objectives for predicting unobserved latent graphs. In addition to its improved performance, \textit{LaGraph} provides explanations for recent successes of predictive models that include invariance-based objectives. We provide theoretical analysis comparing \textit{LaGraph} to related methods in different domains. Our experimental results demonstrate the superiority of \textit{LaGraph} in performance and the robustness to the decreasing training sample size on both graph-level and node-level tasks.
\end{abstract}

\section{Introduction}
Self-supervised learning (SSL) methods seek to use supervisions provided by data itself and design effective pretext learning tasks. These methods allow deep models to learn from a massive amount of unlabeled data and have achieved promising successes in natural language processing~\citep{devlin2019bert, wu2019self, wang2019self} and image tasks~\citep{batson2019noise2self, XieNoise2Same, he2020momentum, chen2020simple}. To use unlabeled graph data, earlier studies~\citep{Perozzi2014deepwalk, grover2016node2vec} adapt sequence-based SSL methods~\citep{Mikolov2013DistributedRO, Mikolov2013EfficientEO} to learn node representations. Inspired by the recent success of SSL in the image domain, a variety of SSL methods based on graph neural networks (GNNs) have been proposed in different learning paradigms. In particular, recent studies~\citep{velikovi2019deep, zhu2020deep, thakoor2021bootstrapped, hassani2020contrastive, you2020graph} construct SSL tasks as unsupervised approaches to learn representations from graph data at either node-level or graph-level; Hu et~al. \yrcite{Hu2020Strategies} propose SSL strategies to pre-train GNNs for downstream tasks; and other studies~\citep{jin2020self, kim2021how} employ SSL as auxiliary tasks to boost the performance of main learning tasks.

Common taxonomies in recent survey works~\citep{xie2022self, liu2020self} consider two categories of SSL methods to train GNNs; namely, contrastive methods and predictive methods. Contrastive methods employ pair-wise discrimination as their pretext learning tasks. It performs transformations or augmentations to obtain multiple views from a graph and trains GNNs to discriminate between jointly sampled view pairs and independently sampled view pairs. In contrast, predictive methods~\citep{hamilton2017inductive, hwang2020self, rong2020grover} train GNNs to predict certain labels obtained from the input graph, such as node reconstruction, connectivity reconstruction, graph statistical properties, and domain knowledge-based targets.

Adapted from the image domain, current state-of-the-art SSL methods for graphs are mostly contrastive. As a drawback, they usually depend on a large training sample size to include a sufficient number of negative samples. With limited computing resources, contrastive methods may not be applicable to large-scale graphs without suffering from performance loss. To address the drawback, BGRL~\citep{thakoor2021bootstrapped} adapts BYOL~\citep{grill2020bootstrap} to the graph domain. BGRL still obtains different views from each given graph, but it eliminates the requirement of negative samples by replacing contrastive objectives with the prediction of offline embedding. BGRL has achieved competitive performance to the contrastive methods. However, unlike contrastive methods grounded by mutual information estimation and maximization, BYOL and BGRL lack theoretical guidance and require implementation measures to prevent collapsing to trivial representations, such as stop gradient, EMA, and normalization layers.

In this work, we propose \textit{LaGraph}, a predictive SSL framework for representation learning of graph data, based on self-supervised latent graph prediction. In particular, we describe the notion of the latent graph and introduce the latent graph prediction as a pretext learning task. We adapt the supervised objective of latent graph prediction into a self-supervised setting by deriving its self-supervised upper bounds, according to which we present the learning framework of \textit{LaGraph}. We provide further justifications of \textit{LaGraph} by comparing it with theoretically sound methods in different domains.
Our experimental results demonstrate the effectiveness of \textit{LaGraph} on both graph-level and node-level representation learning, where a remarkable performance boost is achieved on a majority of datasets with higher stability to smaller batch sizes or training on subsets of nodes. Our code is available under the DIG library~\footnote{\url{https://github.com/divelab/DIG}.}~\cite{JMLR:v22:21-0343}. 

\paragraph{Relations with Prior Work:} Both \textit{LaGraph} and some existing contrastive methods~\citep{you2020graph, zhu2020deep, Hu2020Strategies} apply node masking. While those contrastive methods use node masking as an augmentation to obtain different views for contrast, \textit{LaGraph} employs it for the computation of the invariance term in its predictive objective. In addition, the objective of BGRL has a similar formulation to the invariance regularization term in our objective. The objectives of LaGraph and BGRL are from different grounding and have essential differences in their computing and effects. While the objective of BGRL is designed and engineered as a variant of contrastive methods, the LaGraph objectives are derived as a whole from the latent graph prediction. Our derived theorems associated with \textit{LaGraph} objectives can explain the success of BGRL to some extent \red{and provide guidance on better adopting objectives related to the invariance regularization on graphs}.

\section{Methods}\label{sec:method}
\subsection{Notations and Problem Formulation}
We consider an undirected graph $G=(V, E)$ with a set of attributed nodes $V$ and a set of edges $E$. We formulate the graph data as a tuple of matrices $(\bm A, \bm X)$, where $\bm A\in \mathds{R}^{|V|\times |V|}$ denotes the adjacency matrix and $\bm X\in \mathds{R}^{|V|\times d}$ denotes the node features of dimension $d$. We employ a graph encoder $\mathcal{E}$ based on graph neural networks (GNNs) to encode each node or graph into a corresponding representation. Namely, we compute the node-level representations or node embedding by $\bm H=\mathcal{E}(\bm A, \bm X)\in \mathds{R}^{|V|\times q}$ and the graph-level representation or graph embedding by $\bm z=\mathcal{R}(\bm H)\in \mathds{R}^{1\times q}$, where $q$ denotes the embedding dimension and $\mathcal{R}: \mathds{R}^{|V|\times q}\to\mathds{R}^{1\times q}$ is a readout function.

Self-supervised representation learning is employed to train the graph encoder $\mathcal{E}$ on \red{a set of $K$ graphs $\{G_i\}_{i=1}^K$} without labels from downstream tasks. In particular, we seek to design effective pre-text learning tasks, whose labels are obtained by task designation or from given data, to train the graph encoder $\mathcal{E}$ and produce informative representations for downstream tasks. Depending on the pre-text learning tasks, the encoder $\mathcal{E}$ is usually trained together with some prediction head $\mathcal{D}$ for predictive SSL or a discriminator for contrastive SSL.

\subsection{Latent Graph Prediction}

Our method considers latent graph prediction as a pretext task to train graph neural networks. In this subsection, we introduce the general notion of latent data, followed by its specific definition for graph data, and the construction of the learning task. For any observed data instance $\bm x$, we assume that there exists a corresponding latent data $\bm x_{\mathcal{I}}$, determining the semantic of $\bm x$, such that the latent data $\bm x_{\mathcal{I}}$ is generated from a prior $p(\bm x_{\mathcal{I}})$ and the observed data instance is further generated from a certain distribution conditioned on the latent data, \emph{i.e.}, $p(\bm x|\bm x_{\mathcal{I}})$. The most common case for the pair of observed data and latent data is the noisy data and its clean version.

When it comes to graph data, \red{we consider the case that an observed graph data $G=(\bm A, \bm X)$ is (noisily) generated from its latent graph $G_{\ell}=(\bm A, \bm F)$ with the same node set and edge set, where node feature matrices $\bm X$ and $\bm F$ for the two graphs have the same dimensionality}. We make two assumptions about the graphs without loss of generality. First, we assume that the observed feature vector $\bm x_v$ of each node $v$ in an observed graph is independently generated from a certain distribution conditioned on the corresponding latent graph. In other words, how $\bm x_v$ is generated from the latent feature $\bm f_v$ is not affected by the generation of other observed feature vectors. Second, we assume that the conditional distribution of the observed graph is centered at the latent graph, \textit{i.e.}, $\mathds{E}[\bm X|G_\ell] = \bm F$. The above assumptions are natural when we have little knowledge about the generation process and are commonly used in other types of data such as the non-structural and zero-mean noise in images. In cases where the generation processes of different nodes are related or the distribution is not centered at $F$, we can still consider the related or biased components into the latent feature and therefore have the assumptions satisfied.

As the latent data usually determine the semantic meaning of observed data, we believe the prediction of the latent graph can provide informative supervision for the learning of both graph-level and node-level representations. We are hence interested in constructing the learning task of latent graph prediction. To perform latent graph prediction, it is straightforward to employ a graph neural network $f:\{0,1\}^{|V|\times |V|}\times\mathds{R}^{|V|\times d}\to \mathds{R}^{|V|\times d}$ that takes an observed graph $G=(\bm A, \bm X)$ as inputs and predicts the feature matrix of its latent graph $G_{\mathcal{I}}=(\bm A, \bm F)$. When the ground truth of the latent feature matrix $\bm F$ is known, the learning objective can be designed as
\begin{equation}\label{eq:mse}
f^* = \arg\min_f\mathds{E}\norm{f(\bm A, \bm X)-\bm F}^2.    
\end{equation}
Intuitively, the latent graph prediction can be considered as a generalized task from noisy data reconstruction that predicts the signal from the noisy data with the objective $\arg\min_f\mathds{E}\norm{f(\bm x)-\bm s}^2$, where the mapping from the signal to the noisy data $p(\bm x|\bm s)$ can usually be explicitly modeled and samples of signal (ground truth) can usually be captured. In the data reconstruction case, pairs of $(\bm x, \bm s)$ can be therefore directly captured or synthetically generated given a certain noise model $p(\bm x|\bm s)$. However, when the task is generalized to latent graph prediction, there is a key challenge preventing us from directly applying the prediction task. That is, whereas there are natural supervisions for noisy data reconstruction, the latent graph is not observed and we are unable to explicitly model the mapping from latent graphs to observed graphs, \emph{i.e.}, the conditional distribution $p(G|G_{\mathcal{I}})$.

\subsection{Self-Supervised Upper Bounds for Latent Graph Prediction}
As discussed in the previous subsection, unlike typical noisy data reconstruction tasks, the latent graph is not observed and $p(G|G_{\mathcal{I}})$ cannot be modeled explicitly. This makes it difficult to construct a direct learning task for latent graph prediction using the objective in Equation~(\ref{eq:mse}). We therefore seek to optimize an alternative objective that approximately optimizes the objective in Equation~(\ref{eq:mse}) without requiring the distribution $p(G|G_{\mathcal{I}})$, nor features $\bm F$ of the latent graph. We now introduce the proposed self-supervised objective for latent graph prediction.

We derive our self-supervised objective without involving $\bm F$ by constructing an upper bound of the objective in Equation~(\ref{eq:mse}). Specifically, we let $J\subset \{0,\cdots,|V|-1\}$ be an arbitrary subset of node indices, $J^c$ denote the complement of set $J$, and $\bm X_{J^c}:=\mathds{1}_{J^c}\odot\bm X+\mathds{1}_{J}\odot\bm M$ be the feature matrix with features of nodes in $V_J$ masked, where $\odot$ denotes element-wise multiplication, $\bm M\in \mathds{R}^{|V|\times d}$ denotes a matrix consisting of independent random noise or zeros as masking values, and $\mathds{1}_{J}\in\mathds{R}^{|V|\times d}$ denotes an indicator matrix such that $\mathds{1}_{J}[i,:]=\textbf{1},\forall i\in J$ and $\mathds{1}_{J}[i,:]=\textbf{0},\forall i\notin J$. We describe the self-supervised upper bound in Theorem~\ref{th:bound}, whose proof is provided in Appendix~\ref{sec:proof1}.

\begin{theorem}\label{th:bound}
Consider a graph $G=(\bm A, \bm X)$ and its latent graph $G_{\mathcal{I}}=(\bm A, \bm F)$. We let the variance of any elements in $\bm X$ be bounded by $\sigma^2$ and $J$ be a subset of nodes $V$ in the graph $G$. For any graph neural network $f:\{0,1\}^{|V|\times |V|}\times\mathds{R}^{|V|\times d}\to \mathds{R}^{|V|\times d}$, we have the following inequality
\begin{equation}
\begin{split}
    \mathds{E}_{\bm A,\bm X,\bm F}&\bigg[\norm{f(\bm A, \bm X)-\bm F}^2 +\norm{\bm{X}-\bm F}^2\bigg]\\
    &\hfill\le\mathds{E}_{\bm A, \bm X}\norm{f(\bm{A},\bm X)-\bm{X}}^2 +\\ 
    2\sigma|V|\,\mathds{E}_J&\left[\frac{\mathds{E}_{\bm A, \bm X}\norm{f_J(\bm A, \bm X)-f_J(\bm A, \bm X_{J^c})}^2}{|J|}\right]^{1/2}.
\label{ineq:1}
\end{split}
\end{equation}
\end{theorem}

Intuitively, the first component in the upper bound derived in Theorem~\ref{th:bound} measures the reconstruction error on the feature matrix $\bm X$ of the given observed graph $G$, enforcing the intermediate representations to be informative. The second component controls how much information is accessible from the input feature of a node $v_i$ when reconstructing the feature of $v_i$, by encouraging the output of a node to be invariant to the missing of its features in the input graph. We then call the first component a reconstruction term and the second component an invariance regularization term. \red{Note that the invariance regularization is only computed on masked nodes in contrast to the BGRL objective, based on different theoretical grounding and leading to a different effect. A more detailed discussion is provided in Section 3.}

In tasks of self-supervised representation learning, we are more interested in graph-level or node-level representations than predicted latent graphs. In these cases, we expect the representations also hold the invariance property held by the final outputs. We, therefore, seek to apply the invariance regularization to the representations, since a regularization applied to the output does not necessarily control the information accessibility of representations produced intermediately in the graph neural network. To do so, we separately consider the encoder $\mathcal{E}$ and decoder $\mathcal{D}$ in the graph neural network $f$. We introduce certain assumptions to the decoder network $\mathcal{D}$ and the readout function $\mathcal{R}$, and derive two additional upper bounds for node-level and graph-level representation learning, respectively in the following corollaries. Proofs of the corollaries are provided in Appendix~\ref{sec:proof2}.

\begin{corollary}\label{col:1}
Let $G=(\bm A, \bm X)$ be a given graph, $G_{\mathcal{I}}=(\bm A, \bm F)$ be its latent graph, $\mathcal{E}$ and $\mathcal{D}$ be a graph encoder and a prediction head (decoder) consisting of fully-connected layers. If the prediction head $\mathcal{D}$ is $\ell$-Lipschitz continuous with respect to $l_2$-norm, we further have the following inequality,
\begin{equation}
\begin{split}
    \mathds{E}\big[\norm{\mathcal{D}(\bm H)-\bm F}^2 &+\norm{\bm{X}-\bm F}^2 \big]
    \le \mathds{E}\norm{\mathcal{D}(\bm H)-\bm X}^2 \\+ 2\sigma&|V|\ell\,\mathds{E}_J\left[\frac{\mathds{E}\norm{\bm H_J-\bm H'_J}^2}{|J|}\right]^{1/2},
\end{split}
\end{equation}
where $\bm H=\mathcal{E}(\bm A, \bm X)$ and $\bm H'=\mathcal{E}(\bm A, \bm X_{J^c})$ denote the node embedding of the given graph and the masked graph, respectively, \red{and $\bm H_J:=\bm H[J,:]$ selects rows with indices in $J$.}
\end{corollary}

\begin{corollary}\label{col:2}
Let $G=(\bm A, \bm X)$ be a given graph, $G_{\mathcal{I}}=(\bm A, \bm F)$ be its hidden latent graph, $\mathcal{E}$ be a graph encoder, $\mathcal{R}$ be a readout function satisfying $k$-Bilipschitz continuity with respect to $l_2$-norm, and $\mathcal{D}$ be a prediction head (decoder). If the prediction head $\mathcal{D}$ is $\ell$-Lipschitz continuous with respect to $l_2$-norm, we have the following inequality,
\begin{equation}
\begin{split}
    \mathds{E}\big[\norm{\mathcal{D}(\bm H)-\bm F}^2& +\norm{\bm{X}-\bm F}^2 \big]
    \le \mathds{E}\norm{\mathcal{D}(\bm H)-\bm X}^2 \\ + & 2\sigma|V|k\ell\,\mathds{E}_J\left[\frac{\mathds{E}\norm{\bm z-\bm z'}^2}{|J|}\right]^{1/2},
\end{split}
\end{equation}
where $\bm z=\mathcal{R}(\bm H)$ and $\bm z'=\mathcal{R}(\bm H')$ denote the graph-level representations of the given graph and the masked graph, respectively.
\end{corollary}

We note that the assumptions and restrictions are natural or practically satisfiable. The assumption that the variance of each element in $\bm X$ is bounded by $\sigma$ holds when node features are from $\{0,1\}^d$ or when feature normalization is applied. The $\ell$-Lipschitz continuous property is common for neural networks. And the $k$-Bilipschitz continuity can be satisfied by applying an injective readout function such as global sum pooling, which is commonly used in graph-level tasks.

\subsection{The \textit{LaGraph} Framework}\label{sec:loss}

\begin{figure*}
    \centering
    \includegraphics[width=\textwidth]{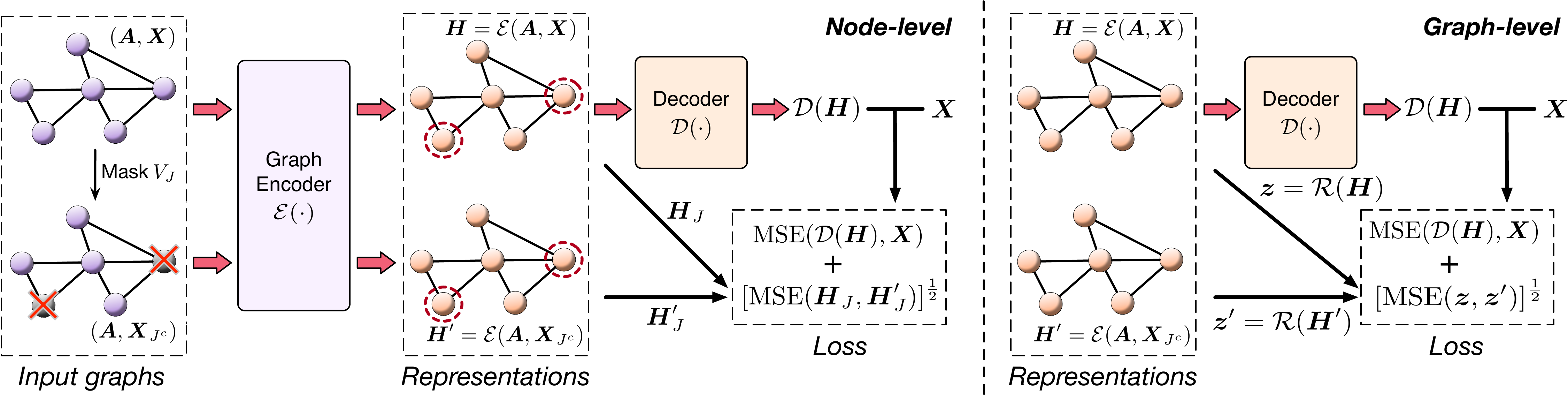}
    \vspace{-10pt}
    \caption{Overview of the \textit{LaGraph} framework. Given a training graph, we randomly mask a small portion $V_J\in V$ of its nodes and input both the original graph and masked graph to the encoder $\mathcal{E}$. \red{Crossed nodes in the figure have all their attributes masked but topology preserved.} The final loss consists of a reconstruction loss on node features and an invariance loss between representations of the original graph and the masked graph. We omit the encoding part of the graph-level framework as frameworks for the two levels mainly differ in whether the invariance term is computed on representations of masked nodes or graph-level representations obtained by $\mathcal{R}$.}
    \label{fig:framework}
\end{figure*}

We design our self-supervised learning framework according to upper bounds derived in Corollary~\ref{col:1} and Corollary~\ref{col:2}. To train encoder $\mathcal{E}$ together with decoder $\mathcal{D}$ under self-supervision, we input to the encoder both the given graph $(\bm A, \bm X)$ and its variation $(\bm A, \bm X_{J^c})$ with a random subset $J$ of node indices for nodes to be masked and obtain node-level representations $\bm H=\mathcal{E}(\bm A, \bm X)$ and $\bm H'=\mathcal{E}(\bm A, \bm X_{J^c})$ for the two graphs respectively. The self-supervised losses are computed on input node features, reconstructed node features, and representations, as demonstrated in Figure~\ref{fig:framework}. 

In particular, we consider a mini-batch of $N$ graphs $\{(\bm A_i, \bm X_i)\}_{i=1}^{N}$ and their corresponding masked variation $\{(\bm A_i, \bm X_{(i, {J_i}^c)})\}_{i=1}^{N}$ where $J_i$ denotes the node indices subset for the $i$-th graph. The self-supervised loss for node-level representation learning follows Corollary~\ref{col:1} and is computed as
\begin{equation}
\begin{split}
    L_{node}(\mathcal{E}, \mathcal{D}&) = \frac{1}{N}\sum_{i=1}^{N}\norm{\mathcal{D}(\bm A_i, \bm H_i)-\bm X_i}^2/{|V_i|} \\+ \alpha&\left[\frac{\sum_{i}\norm{\mathds{1}_{J_i}\odot\bm H_i-\mathds{1}_{J_i}\odot\bm H'_i}^2}{\sum_i|{J_i}|} \right]^{1/2},
\end{split}
\end{equation}
where $\alpha$ is a hyper-parameter corresponding to the multiplier $2\sigma\ell$ in Corollary~\ref{col:1}. To fulfill the conditions in Corollary~\ref{col:1}, we employ fully-connected layers instead of graph convolutional layers in the decoder $\mathcal{D}$.

Similarly, using the same notations above, the self-supervised loss for graph-level representation learning follows Corollary~\ref{col:2} and is computed as
\begin{equation}
\begin{split}
    L_{graph}(\mathcal{E}, \mathcal{D}) = & \frac{1}{N}\sum_{i=1}^{N}\norm{\mathcal{D}(\bm A_i, \bm H_i)-\bm X_i}^2/{|V_i|} \\+ &\alpha'\left[\sum_{i}\norm{\bm z_i-\bm z'_i}^2/\sum_i|{J_i}|\right]^{1/2},
\end{split}
\end{equation}
where $\bm z_i=\mathcal{R}(\bm H_i)$ and $\bm z'_i=\mathcal{R}(\bm H'_i)$ denote the graph-level representations obtained by applying readout function $\mathcal{R}$ to the node-level representations, respectively, and $\alpha'$ is a hyper-parameter corresponding to the multiplier $2\sigma k\ell$ in Corollary~\ref{col:2}. To fulfill the conditions in Corollary~\ref{col:2}, we employ global sum pooling as the readout function $\mathcal{R}$, where as the decoder $\mathcal{D}$ here can consists of either fully-connected layers or graph convolutional layers. 

The pseudo-code for node-level and graph-level objective computations are provided in Algorithm~\ref{alg:1} and Algorithm~\ref{alg:2}, respectively.

\begin{algorithm}[h]
\caption{LaGraph node-level objective}\label{alg:1}
\textbf{Inputs:} A mini-batch of graphs $\{G_1,\cdots,G_N\}$, the encoder $\mathcal{E}$, the prediction head $\mathcal{D}$, and the hyper-parameter $\alpha$. \Comment{$G_i = (\bm A_i, \bm X_i)$}
\begin{algorithmic}
\For{$i$ in $1,\cdots,N$} 
\State Generate random $J_i\in\{0,1\}^{|V_i|\times 1}$, $\bm M\in\mathds{R}^{|V_i|\times d}$
\State $\bm X_{i,J_i^c} \gets \mathds{1}_{J_i^c}\odot\bm X+\mathds{1}_{J_i}\odot\bm M$ \Comment{Randomly mask nodes}
\State $\bm H_i \gets \mathcal{E}(\bm A_i, \bm X_i)$ \Comment{Compute node representations}
\State $\bm H'_i \gets \mathcal{E}(\bm A_i, \bm X_{i,J_i^c})$ 
\State $\bm X_{rec} \gets \mathcal{D}(\bm A_i, \bm H_i)$ \Comment{Reconstructed node attributes}
\State $\ell_{rec, i} \gets \norm{\bm X_{rec,i}-\bm X_i}^2/|V_i|$
\State $\ell_{inv, i} \gets \norm{\mathds{1}_{J_i}\odot\bm H_i - \mathds{1}_{J_i}\odot\bm H'_i}^2$
\EndFor
\State $L(\mathcal{E}, \mathcal{D}; \{G_1,\cdots,G_N\}) = \frac{1}{N}\sum_i \ell_{rec, i} + \alpha(\sum_i\ell_{inv, i}/\sum_i|J_i|)^{1/2}$
\end{algorithmic}
\end{algorithm}

\begin{algorithm}[h]
\caption{LaGraph graph-level objective}\label{alg:2}
\textbf{Inputs:} A mini-batch of graphs $\{G_1,\cdots,G_N\}$, the encoder $\mathcal{E}$, the prediction head $\mathcal{D}$, the readout function $\mathcal{R}$, and the hyper-parameter $\alpha$. \Comment{$G_i = (\bm A_i, \bm X_i)$}
\begin{algorithmic}
\For{$i$ in $1,\cdots,N$} 
\State Generate random $J_i\in\{0,1\}^{|V_i|\times 1}$, $\bm M\in\mathds{R}^{|V_i|\times d}$
\State $\bm X_{i,J_i^c} \gets \mathds{1}_{J_i^c}\odot\bm X+\mathds{1}_{J_i}\odot\bm M$ \Comment{Randomly mask nodes}
\State $\bm H_i \gets \mathcal{E}(\bm A_i, \bm X_i)$ \Comment{Compute node embeddings}
\State $\bm H'_i \gets \mathcal{E}(\bm A_i, \bm X_{i,J_i^c})$ 
\State $\bm z_i \gets \mathcal{R}(\bm H_i)$ \Comment{Readout graph representations}
\State $\bm z'_i \gets \mathcal{R}(\bm H'_i)$
\State $\bm X_{rec} \gets \mathcal{D}(\bm A_i, \bm H_i)$ \Comment{Reconstructed node attributes}
\State $\ell_{rec, i} \gets \norm{\bm X_{rec,i}-\bm X_i}^2/|V_i|$
\State $\ell_{inv, i} \gets \norm{\bm z_i - \bm z'_i}^2$
\EndFor
\State $L(\mathcal{E}, \mathcal{D}; \{G_1,\cdots,G_N\}) = \frac{1}{N}\sum_i \ell_{rec, i} + \alpha(\sum_i\ell_{inv, i}/\sum_i|J_i|)^{1/2}$
\end{algorithmic}
\end{algorithm}

\section{Theoretical Analysis and Relations with Prior Work}
In this section, we further theoretically justify and motivate \textit{LaGraph} by providing comparisons and connections between our method and existing related methods, including denoising autoencoders~\citep{vincent2010stacked, wang2017mgae}, information bottleneck principle~\citep{tishby2000information}, and contrastive methods based on local-global mutual information maximization~\citep{velikovi2019deep, sun2019infograph, hassani2020contrastive}. We also discuss the relation and difference to BGRL~\citep{thakoor2021bootstrapped} and Barlow-Twin~\citep{zbontar2021barlow}.

\subsection{Denoising Autoencoders}
Denoising autoencoders employ an encoder-decoder network architecture and perform self-supervised training by masking or corrupting a portion of dimensions of the given data and reconstructing the masked or corrupted value given their context. Such an approach has been also applied for self-supervised image denoising~\citep{batson2019noise2self}, known as blind-spot denoising. Similar to our method, the denoising autoencoder can be also viewed as an approximation of the latent graph prediction. Using the same notation in Section~\ref{sec:method}, we formulate the connection between latent graph prediction and the graph denoising autoencoder in the following theorem.
\begin{theorem}
Let $J$ be a uniformly sampled subset of node indices of the given graph $(\bm A, \bm X)$, $\mathcal{F}$ be the class of all graph neural networks, and $\mathcal{F}^*$ be the class of graph neural networks such that $f^*_J(\bm A, \bm X)$ does not depend on $\bm X_J$, for any $J$ and $f^*\in\mathcal{F}^*$. Given any graph neural network $f\in\mathcal{F}$, there exist $f^*\in\mathcal{F}^*$ and $f'\in\mathcal{F}$ such that
\begin{align}
    &\mathds{E}_{\bm A, \bm X,\bm F}\left[\norm{f(\bm{A},\bm X)-\bm F}^2 +\norm{\bm{X}-\bm F}^2 \right]\\
    =& \mathds{E}_{\bm A, \bm X}\norm{f(\bm{A},\bm X)-\bm{X}}^2 + \nonumber\\
    &\quad\quad\quad\quad
    \mathds{E}_{\bm A, \bm X, \bm F}\left[2\inner{f(\bm{A},\bm X)-\bm F, \bm X - \bm F}\right]\\
    \approx& \mathds{E}_{\bm A, \bm X}\norm{f^*(\bm{A},\bm X)-\bm{X}}^2\\
    =& |V|\mathds{E}_J\mathds{E}_{\bm A, \bm{X}}\norm{f'_J(\bm A, \bm{X}_{J^c})-\bm{X}_J}^2/|J|.
\end{align}
\end{theorem}
Equation~(7) is proved in the proof of Theorem~1. It can be verified that the second term, \textit{i.e.}, the expectation of the inner product, in Equation~(7) reduces to zero when the neural network $f$ satisfies that $f_J(\bm A, \bm X)$ does not depend on $\bm X_J$, for any $J$, according to Batson and Royer~\yrcite{batson2019noise2self}. The objective can be therefore approximated by Equation~(8) with the neural network $f^*$ satisfying such a property. To let any graph neural network $f$ satisfy the property, one can apply masks to a portion of nodes indexed by $J$ so that their original value is inaccessible by $f$ when predicting $f_J(\bm A, \bm X)$. Therefore, the latent graph prediction objective under supervision can be further approximated by Equation~(9), which describes the objective of a graph denoising autoencoder.

A substantial difference between our method and the denoising autoencoder lies in how to handle the inner product term in Equation~(7). In particular, the denoising autoencoder forces the term to be zero by assuming certain properties of the graph neural network, whereas our method derives an upper bound, \textit{i.e.}, the invariance term, for the inner product. Theoretically, the graph denoising autoencoder is equivalent to our framework with an infinite weight scalar for the invariance term. As a drawback, when $f_J(\bm{A},\bm X)$ does not depend on $\bm X_J$, the learned representations can be less informative as representations of nodes in $V_J$ do not include the information of $\bm X_J$, for any $J$, leading to performance loss. Our proposed upper bounds allow an encoder to access a certain level of information of the masked nodes, whose representations can be as good as ones from supervised learning. In fact, our method can be viewed as an autoencoder with an invariance regularization.

\subsection{The Information Bottleneck Principle}
The information bottleneck principle~\citep{tishby2000information} is a technique for data compression and signal processing in the field of information theory, and has been widely applied in deep learning problems~\citep{tishby2015deep, michael2018on}. Let $\bm X$ be a random variable to be compressed, $\tilde{\bm X}$ be an observed relevant variable, and $\bm Z$ denote the compressed representation of $\bm X$. The information bottleneck principle seeks to optimize the following problem
\begin{equation}\label{eq:ibn}
    \bm Z^* = \arg\min_{\bm Z} I(\bm Z;\tilde{\bm X}) - \beta I(\bm Z;\bm X),
\end{equation}
where $I(\cdot;\cdot)$ denotes the mutual information and $\beta>1$ is a Lagrange multiplier. The work Barlow Twin~\citep{zbontar2021barlow} has discussed a connection between the information bottleneck principle and self-supervised learning. In particular, to apply information bottleneck to SSL, one usually obtain $\tilde{\bm X}$ by performing augmentations or distortions on the given data $\bm X$. And Equation~(\ref{eq:ibn}) can be rewritten into
\begin{align}
    \bm Z^* 
    = \arg\min_{\bm Z} \big[H(\bm Z)-&H(\bm Z|\tilde{\bm X})\big]\\& - \beta \big[H(\bm Z)-H(\bm Z|\bm X)\big]\\
    = \arg\min_{\bm Z} H(\bm Z|\bm X)& - \lambda H(\bm Z),\label{eq:12}
\end{align}
where $\lambda=\frac{\beta-1}{\beta}>0$ is a weight scalar. Intuitively, the conditional entropy $H(\bm Z|\bm X)$ is to be minimized, indicating that the distortion should add no additional information to the representation $\bm Z$. In other words, the representation $\bm Z$ should be as invariant as possible to distortions applied to $\bm X$. In addition, the entropy $H(\bm Z)$ is to be maximized, indicating that the representation $\bm Z$ itself should be as informative as possible.

The two terms in objectives of \textit{LaGraph} correspond to the terms in Equation~(\ref{eq:12}). In particular, the invariance term corresponding to $H(\bm Z|\bm X)$ and the reconstruction term aims to ensure informative representations, \textit{i.e.}, to maximize $H(\bm Z)$. Objectives in existing SSL methods such as BYOL~\citep{grill2020bootstrap}, its variation BGRL~\citep{thakoor2021bootstrapped} in graph domain, and Barlow Twin~\citep{zbontar2021barlow} also include invariance terms corresponding to $H(\bm Z|\bm X)$. To encourage informative representations, Barlow Twin further includes a redundancy reduction term to minimize the cross-correlation between different dimensions of the representation, as a proxy of the maximization of $H(\bm Z)$. In addition, the InfoNCE (NT-XENT) loss employed in some contrastive learning methods~\citep{you2020graph, zhu2020deep} induces a similar effect, according to Zbontar et~al.~\yrcite{zbontar2021barlow}. Both Equation~(\ref{eq:12}) and the derivation of \textit{LaGraph} objectives indicate the importance of the invariance term in SSL objectives. In addition, compared to the redundancy reduction term in Barlow Twin and the noise contrast in InfoNCE, \textit{LaGraph} objectives can directly guarantee the learning of informative representations measured by the reconstruction capability. 

\subsection{Contrastive Learning by Maximizing Local-Global Mutual Information}
Motivated by Deep InfoMax~\citep{hjelm2018learning}, recent graph self-supervised learning methods~\citep{velikovi2019deep, sun2019infograph, hassani2020contrastive} constructs their learning tasks by maximizing the mutual information between local (node-level) representations and a global (graph-level) summary of the graph. Practically, as a $k$-layer encoder $\mathcal{E}$ has the receptive field of at most $k$-hop neighborhood, the goal becomes the maximization of the mutual information between local representations and their $k$-hop neighborhood, formulated as
\begin{equation}
    \mathcal{E}^* = \arg\max_{\mathcal{E}} \sum_{i=1}^{|V|}I(\bm X_i^{(k)}; \mathcal{E}_i(\bm A, \bm X)),
\end{equation}
where $I$ denotes the mutual information, $\bm X_i^{(k)}$ is the $k$-hop neighborhood of node $i$, $\mathcal{E}$ is a graph encoder with $k$ GNN layers, and $\mathcal{E}_i(\bm A, \bm X)$ denotes the local representation of node $i$. The learning objective is motivated by the goal that the local representations should contain as much the global information of the entire graph (or the $k$-hop neighborhood) as possible. 

As for \textit{LaGraph}, the reconstruction term encourages representations to contain sufficient information to reconstruct the input features while the invariance term limits the information accessibility from a local node when reconstructing its features. The two terms in the objective jointly promote node representations to learn limited local information and as much contextual information from the neighborhood as possible for reconstruction. It hence has a similar effect to the local-global mutual information maximization.

\subsection{Other Invariance-Based Objectives}\label{sec:vsbgrl}

Recent self-supervised learning objectives such as BGRL, Barlow-Twin, and the consistency regularization~\citep{wei2021theoretical} have similar invariance terms as one in the LaGraph objective. Specifically, BGRL minimizes the difference between representations of two augmented views. In spite of the similarity, the invariance terms in LaGraph and other objectives have different grounding and effects.

Regarding how the objectives are computed, the invariance term in the LaGraph objective for node-level representation learning is computed only on masked nodes, in contrast to BGRL and Barlow-Twins objectives where invariance of all nodes are computed. It is worth noting that the proposed objective is an upper bound to the latent graph prediction only if the invariance is computed on the masked nodes, according to the derivation in the proof of Theorem 1. Intuitively, during the computation of a node representation, the invariance term in LaGraph enforces the encoder to capture less information from the node itself and more contextual information. Computing the invariance regularization term on unmasked nodes could lead to a contradicted effect, i.e., discouraging encoders to capture information from contextual nodes, as it lets the representation remain consistent when its masked neighbor nodes are changed. We believe the derivation and the intuition of the proposed objective can provide insights on adopting the invariance regularization into graph self-supervised learning studies.

\section{Experiments}
We conduct experiments on both node-level and graph-level self-supervised representation learning tasks with datasets used in two most recent state-of-the-art methods for SSL~\citep{you2020graph, thakoor2021bootstrapped}. For graph-level tasks, we follow GraphCL~\citep{you2020graph} to perform evaluations on eight graph classification datasets~\citep{NCI1,proteins,DD,mutag,yanardag2015deep} from TUDataset~\citep{Morris2020tudataset}. For node-level tasks, as the citation network datasets~\citep{mccallum2000cora, citeseer, sen2008pubmed} are recognized to be saturated and unreliable for GNN evaluation~\citep{shchur2018pitfalls, thakoor2021bootstrapped}, we follow Thakoor et~al.~\yrcite{thakoor2021bootstrapped} to include four transductive node classification datasets from Shchur et~al.~\yrcite{shchur2018pitfalls}, including Amazon Computers, Amazon Photos from the Amazon Co-purchase Graph~\citep{mcauley2015image}, Coauthor CS, and Coauthor Physics from the Microsoft Academic Graph~\citep{sinha2015overview}. We further include three larger-scale inductive datasets, PPI, Reddit, and Flickr, for node-level classification used in SUBG-CON~\citep{jiao2020sub}. 

We follow You et~al.~\yrcite{you2020graph} and Zhu et~al.~\yrcite{zhu2020deep} for the standard linear evaluation protocols at graph-level and node-level, respectively. In particular, for both levels, we first train the graph encoder on unlabeled graph datasets with the corresponding self-supervised objective. We then compute and freeze the corresponding representations and train a linear classification model on top of the fixed representations with their corresponding labels. Linear SVM and the regularized logistic regression are employed as linear classifiers for graph-level datasets and node-level datasets, according to You et~al.~\yrcite{you2020graph} and Zhu et~al.~\yrcite{zhu2020deep}, respectively. For inductive node-level datasets, the self-supervised training is only performed on graphs in the training datasets whereas the test graphs are unavailable during the self-supervised training.

\begin{table*}[t]
    \small
    \centering
    \caption{Performance on graph-level classification tasks, scores are averaged over 5 runs. Bold and underlined numbers highlight the top-2 performance. OOM indicates running out-of-memory on a 56GB Nvidia A6000 GPU.}\label{tab:graph}
    \begin{tabular}{ccccccccc}
    \toprule
                      & NCI1       & PROTEINS   & DD         & MUTAG       & COLLAB     & RDT-B      & RDT-M5K    & IMDB-B     \\ 
    \hline
    GL                & --       & --       & --         & 81.7±2.1  & --         & 77.3±0.2 & 41.0±0.2 & 65.9±1.0 \\
    WL                & 80.0±0.5 & 72.9±0.6 & --         & 80.7±3.0  & --         & 68.8±0.4 & 46.1±0.2 & 72.3±3.4 \\
    DGK               & 80.3±0.5 & 73.3±0.8 & --         & 87.4±2.7  & --         & 78.0±0.4 & 41.3±0.2 & 67.0±0.6 \\
    \hline
    Node2Vec          & 54.9±1.6 & 57.5±3.6 & 75.1±0.5   & 72.6±10.2 & 55.7±0.2   & 73.8±0.5 & 34.1±0.4 & 50.0±0.8   \\
    Sub2Vec           & 52.8±1.5 & 53.0±5.6 & 73.6±1.5   & 61.1±15.8 & 62.1±1.4   & 71.5±0.4 & 36.7±0.4 & 55.3±1.5 \\
    Graph2Vec         & 73.2±1.8 & 73.3±2.1 & 76.2±0.1   & 83.2±9.3  & 59.9±0.0   & 75.8±1.0 & 47.9±0.3 & 71.1±0.5 \\
    GAE               & 73.3±0.6 & 74.1±0.5 & 77.9±0.5 	 & 84.0±0.6  & 56.3±0.1   & 74.8±0.2 & 37.6±1.6 & 52.1±0.2 \\
    VGAE              & 73.7±0.3 & 74.0±0.5 & 77.6±0.4 	 & 84.4±0.6  & 56.3±0.0   & 74.8±0.2 & 39.1±1.6 & 52.1±0.2 \\
    \hline
    InfoGraph         & 76.2±1.1 & \underline{74.4±0.3} & 72.9±1.8 & 89.0±1.1  & 70.7±1.1 & 82.5±1.4 & 53.5±1.0 & 73.0±0.9 \\
    GraphCL           & \underline{77.9±0.4} & 74.4±0.5 & \textbf{78.6±0.4} & 86.8±1.3  & 71.4±1.2 & \underline{89.5±0.8} & \underline{56.0±0.3} & 71.1±0.4 \\
    MVGRL             & 75.1±0.5          & 71.5±0.3          & OOM          & \underline{89.7±1.1}    & OOM          & 84.5±0.6   & OOM          & \textbf{74.2±0.7}   \\ 
    \hline
    LaGraph              & \textbf{79.9±0.5} & \textbf{75.2±0.4} & \underline{78.1±0.4} & \textbf{90.2±1.1}  & \textbf{77.6±0.2} & \textbf{90.4±0.8} & \textbf{56.4±0.4} & \underline{73.7±0.9} \\ 
    \bottomrule
    \end{tabular}
    \label{tab:graph_lv}
\end{table*}

\subsection{Comparisons with Baselines}
We perform experiments on both graph-level and node-level datasets to demonstrate the effectiveness of \textit{LaGraph}. We construct our model and losses according to Section~\ref{sec:loss}. Detailed model configurations, training settings, and dataset statistics are provided in Appendix~\ref{sec:config}.

\paragraph{Graph-level Datasets.}
We evaluate the performance of \textit{LaGraph} in terms of the linear classification accuracy and compare it with three kernel-based methods including graphlet kernel (GL)~\citep{shervashidze2009efficient}, Weisfeiler-Lehman kernel (WL)~\citep{shervashidze2011weisfeiler}, and deep graph kernel (DGK)~\citep{yanardag2015deep}, together with five unsupervised methods including Node2Vec~\citep{grover2016node2vec}, Sub2Vec~\citep{adhikari2018sub2vec}, Graph2Vec~\citep{narayanan2017graph2vec}, GAE and VGAE~\citep{kipf2016variational}. We further compare the results with recent SOTA SSL methods based on contrastive learning, including InfoGraph~\citep{sun2019infograph} , MVGRL~\citep{hassani2020contrastive}, and GraphCL~\citep{you2020graph}. Results in Table~\ref{tab:graph} show that \textit{LaGraph} outperforms the current SOTA methods on a majority of datasets and is on par with the best performance on the rest of datasets. Additional results adopting \textit{LaGraph} as a pre-training strategy under the semi-supervised learning setting are provided in Appendix~\ref{sec:semi}.

\paragraph{Node-level Datasets.} 
We perform node-level experiments on both transductive and inductive learning tasks. Transductive self-supervised learning of node representation allows utilization of all data at hand to pre-train GNNs for downstream tasks. Although labels of nodes are not visible during pre-training, patterns and information present in all nodes are observed. In contrast to transductive learning, inductive self-supervised learning only allows using a portion of data to pre-train GNNs, while holding out a certain amount of data for downstream tasks. Our inductive tasks include two cases. First, the PPI dataset consists of 24 graphs, and the training and testing nodes are split by graphs. In this case, the inductive task is considered across multiple graphs. In other words, node representations are learned from training graphs, and the encoder is evaluated on testing graphs. Second, Flickr and Reddit each consist of only one graph, the training and testing nodes are from the same graph. During self-supervised training, all test nodes are masked-out. During evaluation, all training nodes are masked-out, i.e., test nodes are unseen nodes of the graph during train. For both cases of inductive learning, data used during the self-supervised training stage and data used during evaluation stage are distinct, but the feature dimensionality should be the same for data used in both stages.

For the evaluation of transductive learning, we compare the performance of \textit{LaGraph} in terms of linear classification accuracy with DeepWalk~\citep{deepwalk}, GAE, VGAE, and six contrastive learning methods including Deep Graph InfoMax (DGI)~\citep{velikovi2019deep}, GMI~\citep{peng2020graph}, MVGRL~\citep{hassani2020contrastive}, GRACE~\citep{zhu2020deep}, GCA~\citep{zhu2021graph}, and BGRL~\citep{thakoor2021bootstrapped}, where BGRL is the current state-of-the-art SSL method for node-level representation learning. We further include the results of directly performing linear classification on raw node features (raw features)
and by supervised training for references. To be consistent with Thakoor et~al.~\yrcite{thakoor2021bootstrapped}, we have ensured that the GPU memory consumption of \textit{LaGraph} is under 16GB for the four transductive datasets. We then perform additional experiments on the larger-scale inductive datasets~\citep{zitnik2017predicting, zeng2019graphsaint,hamilton2017inductive} and compare our results in terms of micro-averaged F1-score with DeepWalk, unsupervised GraphSAGE~\citep{hamilton2017inductive}, DGI, GMI, SUBG-CON~\citep{jiao2020sub} and BGRL. Results for both transductive datasets and inductive datasets shown in Table~\ref{tab:node}. As there is no official BGRL implementation available at the time our experiments are conducted, results with $*$ are obtained from an unofficial public implementation\footnote{\url{https://github.com/namkyeong/bgrl_pytorch}.}. Results suggest competitive performance of \textit{LaGraph} compared to the existing SOTA methods. Moreover, LaGraph consumes even less memory than BGRL, which requires twice the memory for its GNN encoders for the EMA parameter update.

\paragraph{Experiment Environment Details.} We train graph-level datasets on a 11GB GeForce RTX 2080 Ti GPU, and node-level datasets on a 56GB Nvidia RTX A6000 GPU. Our experiments are implemented with PyTorch 1.7.0 and PyTorch Geometric 1.7.0. All neural networks employ batch normalization~\citep{ioffe2015batch}, and are optimized with Adam optimizer~\citep{kingma2014adam}. We initialize GNNs with Xavier initialization~\citep{XavierInit}.

\begin{table*}
    \small
    \centering
    \caption{Performance on node-level datasets, 20 runs averaged. Results of SSL methods with the best performance are highlighted in bold numbers. \textit{Left}: Mean classification accuracy on transductive datasets, with baseline results from Thakoor et~al.~\yrcite{thakoor2021bootstrapped}.
    \textit{Right}: Micro-averaged F1 scores on larger-scale inductive datasets, with baseline results from Thakoor et~al.~\yrcite{thakoor2021bootstrapped} and Jiao et~al.~\yrcite{jiao2020sub}. 
    }\label{tab:node}
        \begin{tabular}{ccccc}
        \toprule
        Transductive           & Am.Comp. & Am.Pht.  & Co.CS & Co.Phy \\
        \hline
        Raw features           & 73.8±0.0   & 78.5±0.0 & 90.4±0.0 & 93.6±0.0  \\
        DeepWalk               & 85.7±0.1   & 89.4±0.1 & 84.6±0.2 & 91.8±0.2  \\
        GAE                    & 87.7±0.3 	& 92.7±0.3 & 92.4±0.2 & 95.3±0.1 \\
        VGAE                   & 88.1±0.3 	& 92.8±0.3 & 92.5±0.2 & 95.3±0.1 \\
        Supervised             & 86.5±0.5   & 92.4±0.2 & 93.0±0.3 & 95.7±0.2  \\
        \hline
        DGI                    & 84.0±0.5   & 91.6±0.2 & 92.2±0.6 & 94.5±0.5  \\
        GMI                    & 82.2±0.3   & 90.7±0.2 & OOM        & OOM         \\
        MVGRL                  & 87.5±0.1   & 91.7±0.1 & 92.1±0.1 & 95.3±0.0  \\
        GRACE                  & 87.5±0.2   & 92.2±0.2 & 92.9±0.0 & 95.3±0.0  \\
        GCA                    & 88.9±0.2   & 92.5±0.2 & 93.1±0.0 & 95.7±0.0  \\
        BGRL                   & \textbf{89.7±0.3}   & 92.9±0.3 & 93.2±0.2 & 95.6±0.1  \\
        \hline
        LaGraph                   & 88.0±0.3        & \textbf{93.5±0.4}      & \textbf{93.3±0.2}      & \textbf{95.8±0.1}       \\
        \bottomrule
        \end{tabular}
    \quad
        \addtolength{\tabcolsep}{-1.2pt}
        \begin{tabular}{cccc}
        \toprule
        Inductive        & PPI      & Flickr   & Reddit          \\
        \hline
        Raw feat.        & 42.5±0.3 & 20.3±0.2 & 58.5±0.1        \\
        GAE              & 75.7±0.0 & 50.7±0.2 & OOM             \\
        VGAE             & 75.8±0.0 & 50.4±0.2 & OOM             \\
        Super-GCN        & 51.5±0.6 & 48.7±0.3 & 93.3±0.1        \\
        Super-GAT        & 97.3±0.2 & OOM      & OOM             \\
        \hline
        GraphSAGE        & 46.5±0.7 & 36.5±1.0 & 90.8±1.1        \\
        DGI              & 63.8±0.2 & 42.9±0.1 & 94.0±0.1        \\
        GMI              & 65.0±0.0 & 44.5±0.2 & 95.0±0.0        \\
        SUBG-CON         & 66.9±0.2 & 48.8±0.1 & \textbf{95.2±0.0}        \\
        BGRL-GCN         & 69.6±0.2 & 50.0±0.3* & OOM*            \\
        BGRL-GAT         & 70.5±0.1 & 44.2±0.1* & OOM*            \\
        \hline
        LaGraph             & \textbf{74.6±0.0}    & \textbf{51.3±0.1}     & \textbf{95.2±0.0} \\
        \bottomrule
        \end{tabular}
\end{table*}

\subsection{Ablation Study}
We further conduct three ablation studies to explore model robustness to smaller batch sizes on graph-level data and to the training with sub-graphs on large-scale node-level datasets. An additional ablation study on the effect of optimizing different objectives is provided in Appendix~\ref{sec:ablat}.

\begin{table*}
    \centering
    \small
    \caption{Model performance when trained on a subset of nodes.}\label{tab:subgraphs}
    \begin{tabular}{clcccccc}
    \toprule
           & \# nodes sampled               & 100    & 1,000   & 2,500   & 5,000    & 10,000   & all      \\
    \hline
           & \% nodes sampled               & 0.22\% & 2.24\% & 5.60\% & 11.20\% & 22.41\% & 100.00\% \\
           & F1-score - \textit{LaGraph}    & 6.07   & 51.12  & 51.12  & 51.27   & 51.29   & 51.26    \\
    Flickr & Memory - \textit{LaGraph}      & 1389MB & 1465MB & 1553MB & 1725MB  & 2065MB  & 4211MB   \\
           & F1-score - GraphCL             & 45.27  & 45.27  & 45.27  & 45.38 	 & 45.45   & 45.48    \\
           & Memory - GraphCL               & 1647MB & 2599MB & 4137MB & 6741MB  & 11905MB & 47939MB   \\
    \hline
           & \% nodes sampled               & 0.07\% & 0.65\% & 1.63\% & 3.25\%  & 6.50\%  & 100.00\% \\
           & F1-score - \textit{LaGraph}    & 5.76   & 95.05  & 95.06  & 95.08   & 95.09   & 95.22    \\
    Reddit & Memory - \textit{LaGraph}      & 1403MB & 1475MB & 1585MB & 1783MB  & 2161MB  & 16933MB   \\
           & F1-score - GraphCL             & 93.24  & 93.24  & 93.25  & 93.31 	 & 93.32    & OOM    \\
           & Memory - GraphCL               & 4199MB & 6117MB & 6687MB & 9297MB  & 14495MB & OOM   \\
    \bottomrule
    \end{tabular}
    \label{tab:sample}
\end{table*}

\begin{figure}
    \centering
    \includegraphics[width=0.48\textwidth]{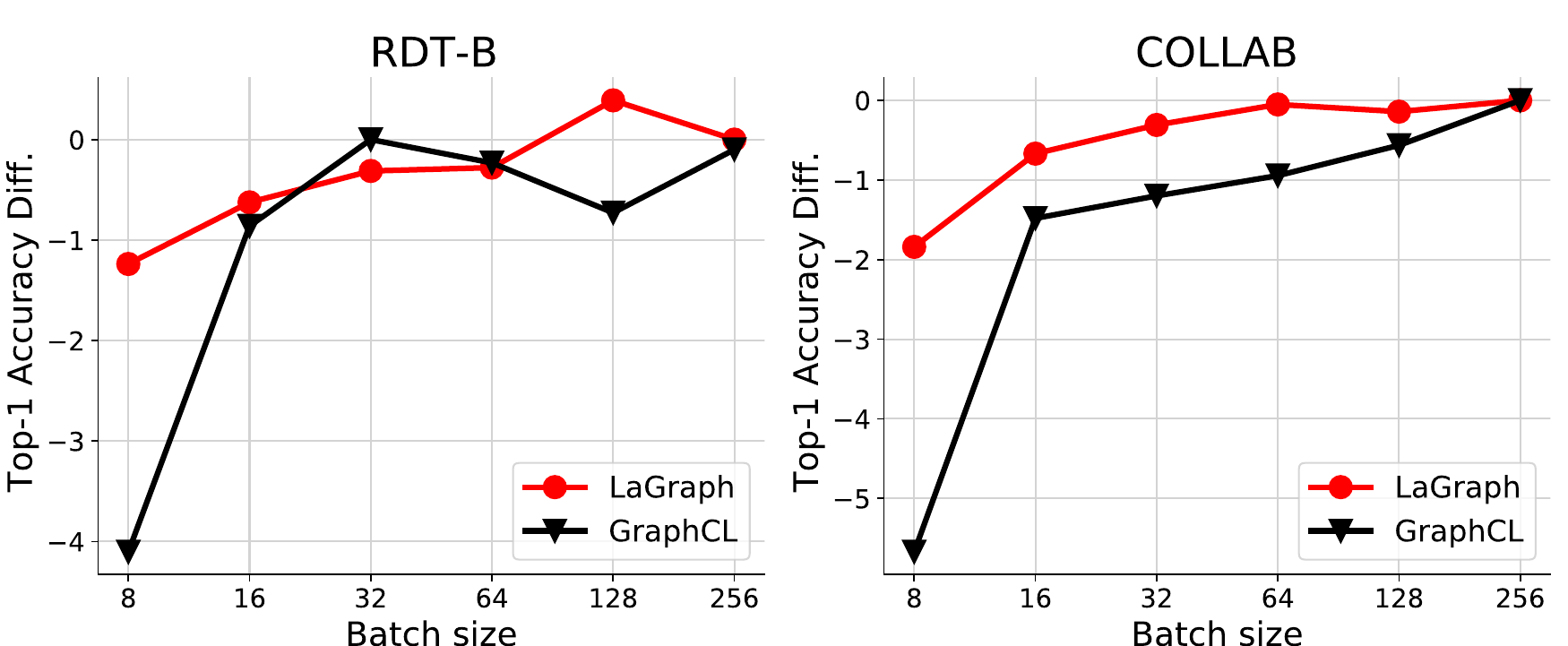}
    \vspace{-12pt}
    \caption{Model robustness to small batch sizes on RDT-B and COLLAB. Shown are relative changes in accuracy over different batch sizes compared to the batch size of 256.}\label{fig:batchsize}
\end{figure}

\paragraph{Robustness to Batch Sizes.}
Different from contrastive learning methods, \textit{LaGraph} does not require negative samples to perform noise contrast or pair-wise discrimination. Therefore, an advantage of \textit{LaGraph} is that the performance is robust to the batch size as it does not depend on large batch sizes with sufficient negative samples. To verify the statement, we perform an ablation study on how model performance changes when decreasing the batch size from 128 to 8 for graph-level datasets. We include corresponding results of GraphCL which uses InfoNCE for references and show the comparisons in Figure~\ref{fig:batchsize}. The results indicate while contrastive methods based on InfoNCE suffer from significant performance loss with a small batch size, \textit{LaGraph} are more robust to the batch size.

\paragraph{Training on Sub-graphs for Large-scale Datasets.}
Training graph encoders on all nodes for some large-scale graphs can be heavily expensive in computation. We hence conduct an ablation study on how training graph encoders on a portion of sampled nodes instead of the entire graph affects the effectiveness of training. Results in Table~\ref{tab:subgraphs} suggest that the model performance remains stable when decreasing the number of nodes until the number becomes extremely small. The collapse is due to the very sparse connectivity and \textit{LaGraph} fails to reconstruct a node from its neighbor nodes as there are no neighbors at all. In contrast, though GraphCL does not collapse at extremely small subsets, it suffers more from performance loss above 1,000 nodes and consumes significantly more GPU memory.

\section{Conclusions and Future Directions}
We introduced \textit{LaGraph}, a state-of-the-art predictive SSL framework whose objectives are based on self-supervised latent graph prediction. We provided theoretical analysis and discussed the relationship between \textit{LaGraph} and theories in different related domains. Experimental results demonstrate the strong effectiveness of the proposed framework and the stability to the training scale for both graph-level and node-level tasks. Currently, our framework mainly considers the latent graph regarding its node features. Further investigation into a latent graph prediction framework that includes richer information such as edge features and latent connectivity into self-supervision can potentially bring additional improvement to the performance. 
We discuss more future directions in Appendix~\ref{sec:fut}.

\section*{Acknowledgments}
This work was supported in part by National Science Foundation grant IIS-2006861.

\bibliographystyle{icml2022}
\bibliography{dive, reference}

\begin{thebibliography}{66}
\providecommand{\natexlab}[1]{#1}
\providecommand{\url}[1]{\texttt{#1}}
\expandafter\ifx\csname urlstyle\endcsname\relax
  \providecommand{\doi}[1]{doi: #1}\else
  \providecommand{\doi}{doi: \begingroup \urlstyle{rm}\Url}\fi

\bibitem[Adhikari et~al.(2018)Adhikari, Zhang, Ramakrishnan, and
  Prakash]{adhikari2018sub2vec}
Adhikari, B., Zhang, Y., Ramakrishnan, N., and Prakash, B.~A.
\newblock {Sub2Vec}: Feature learning for subgraphs.
\newblock In \emph{Pacific-Asia Conference on Knowledge Discovery and Data
  Mining}, pp.\  170--182. Springer, 2018.

\bibitem[Batson \& Royer(2019)Batson and Royer]{batson2019noise2self}
Batson, J. and Royer, L.
\newblock Noise2{S}elf: Blind denoising by self-supervision.
\newblock In \emph{Proceedings of the 36th International Conference on Machine
  Learning}, volume~97, pp.\  524--533, 2019.

\bibitem[Borgwardt et~al.(2005)Borgwardt, Ong, Schönauer, Vishwanathan, Smola,
  and Kriegel]{proteins}
Borgwardt, K.~M., Ong, C.~S., Schönauer, S., Vishwanathan, S. V.~N., Smola,
  A.~J., and Kriegel, H.-P.
\newblock Protein function prediction via graph kernels.
\newblock \emph{Bioinformatics}, 21:\penalty0 i47--i56, 06 2005.
\newblock ISSN 1367-4803.

\bibitem[Chen et~al.(2020)Chen, Kornblith, Norouzi, and Hinton]{chen2020simple}
Chen, T., Kornblith, S., Norouzi, M., and Hinton, G.
\newblock A simple framework for contrastive learning of visual
  representations.
\newblock In \emph{International conference on machine learning}, pp.\
  1597--1607. PMLR, 2020.

\bibitem[Debnath et~al.(1991)Debnath, Lopez~de Compadre, Debnath, Shusterman,
  and Hansch]{mutag}
Debnath, A.~K., Lopez~de Compadre, R.~L., Debnath, G., Shusterman, A.~J., and
  Hansch, C.
\newblock Structure-activity relationship of mutagenic aromatic and
  heteroaromatic nitro compounds correlation with molecular orbital energies
  and hydrophobicity.
\newblock \emph{Journal of Medicinal Chemistry}, 34\penalty0 (2):\penalty0
  786--797, 02 1991.

\bibitem[Devlin et~al.(2019)Devlin, Chang, Lee, and Toutanova]{devlin2019bert}
Devlin, J., Chang, M., Lee, K., and Toutanova, K.
\newblock {BERT:} pre-training of deep bidirectional transformers for language
  understanding.
\newblock In \emph{Proceedings of the 2019 Conference of the North American
  Chapter of the Association for Computational Linguistics: Human Language
  Technologies}, pp.\  4171--4186, 2019.

\bibitem[Dobson \& Doig(2003)Dobson and Doig]{DD}
Dobson, P.~D. and Doig, A.~J.
\newblock Distinguishing enzyme structures from non-enzymes without alignments.
\newblock \emph{Journal of Molecular Biology}, 330\penalty0 (4):\penalty0
  771--783, 2003.
\newblock ISSN 0022-2836.

\bibitem[Giles et~al.(1998)Giles, Bollacker, and Lawrence]{citeseer}
Giles, C.~L., Bollacker, K.~D., and Lawrence, S.
\newblock Citeseer: An automatic citation indexing system.
\newblock In \emph{Proceedings of the Third ACM Conference on Digital
  Libraries}, pp.\  89–98. Association for Computing Machinery, 1998.
\newblock ISBN 0897919653.

\bibitem[Glorot \& Bengio(2010)Glorot and Bengio]{XavierInit}
Glorot, X. and Bengio, Y.
\newblock Understanding the difficulty of training deep feedforward neural
  networks.
\newblock In \emph{Proceedings of the Thirteenth International Conference on
  Artificial Intelligence and Statistics}, volume~9, pp.\  249--256, 2010.

\bibitem[Grill et~al.(2020)Grill, Strub, Altch\'{e}, Tallec, Richemond,
  Buchatskaya, Doersch, Avila~Pires, Guo, Gheshlaghi~Azar, Piot, kavukcuoglu,
  Munos, and Valko]{grill2020bootstrap}
Grill, J.-B., Strub, F., Altch\'{e}, F., Tallec, C., Richemond, P.,
  Buchatskaya, E., Doersch, C., Avila~Pires, B., Guo, Z., Gheshlaghi~Azar, M.,
  Piot, B., kavukcuoglu, k., Munos, R., and Valko, M.
\newblock Bootstrap your own latent: A new approach to self-supervised
  learning.
\newblock In \emph{Advances in Neural Information Processing Systems},
  volume~33, pp.\  21271--21284, 2020.

\bibitem[Grover \& Leskovec(2016)Grover and Leskovec]{grover2016node2vec}
Grover, A. and Leskovec, J.
\newblock node2vec: Scalable feature learning for networks.
\newblock In \emph{Proceedings of the 22nd ACM SIGKDD international conference
  on Knowledge discovery and data mining}, pp.\  855--864, 2016.

\bibitem[Hamilton et~al.(2017)Hamilton, Ying, and
  Leskovec]{hamilton2017inductive}
Hamilton, W., Ying, Z., and Leskovec, J.
\newblock Inductive representation learning on large graphs.
\newblock In \emph{Advances in Neural Information Processing Systems}, pp.\
  1024--1034, 2017.

\bibitem[Hassani \& Khasahmadi(2020)Hassani and
  Khasahmadi]{hassani2020contrastive}
Hassani, K. and Khasahmadi, A.~H.
\newblock Contrastive multi-view representation learning on graphs.
\newblock In \emph{International Conference on Machine Learning}, pp.\
  4116--4126. PMLR, 2020.

\bibitem[He et~al.(2020)He, Fan, Wu, Xie, and Girshick]{he2020momentum}
He, K., Fan, H., Wu, Y., Xie, S., and Girshick, R.
\newblock Momentum contrast for unsupervised visual representation learning.
\newblock In \emph{Proceedings of the IEEE/CVF conference on computer vision
  and pattern recognition}, pp.\  9729--9738, 2020.

\bibitem[Hjelm et~al.(2019)Hjelm, Fedorov, Lavoie-Marchildon, Grewal, Bachman,
  Trischler, and Bengio]{hjelm2018learning}
Hjelm, R.~D., Fedorov, A., Lavoie-Marchildon, S., Grewal, K., Bachman, P.,
  Trischler, A., and Bengio, Y.
\newblock Learning deep representations by mutual information estimation and
  maximization.
\newblock In \emph{International Conference on Learning Representations}, 2019.

\bibitem[Hu et~al.(2020)Hu, Liu, Gomes, Zitnik, Liang, Pande, and
  Leskovec]{Hu2020Strategies}
Hu, W., Liu, B., Gomes, J., Zitnik, M., Liang, P., Pande, V., and Leskovec, J.
\newblock Strategies for pre-training graph neural networks.
\newblock In \emph{International Conference on Learning Representations}, 2020.

\bibitem[Hwang et~al.(2020)Hwang, Park, Kwon, Kim, Ha, and Kim]{hwang2020self}
Hwang, D., Park, J., Kwon, S., Kim, K., Ha, J.-W., and Kim, H.~J.
\newblock Self-supervised auxiliary learning with meta-paths for heterogeneous
  graphs.
\newblock In \emph{Advances in Neural Information Processing Systems}, pp.\
  10294--10305, 2020.

\bibitem[Ioffe \& Szegedy(2015)Ioffe and Szegedy]{ioffe2015batch}
Ioffe, S. and Szegedy, C.
\newblock Batch normalization: Accelerating deep network training by reducing
  internal covariate shift.
\newblock In \emph{International conference on machine learning}, pp.\
  448--456. PMLR, 2015.

\bibitem[Jiao et~al.(2020)Jiao, Xiong, Zhang, Zhang, Zhang, and
  Zhu]{jiao2020sub}
Jiao, Y., Xiong, Y., Zhang, J., Zhang, Y., Zhang, T., and Zhu, Y.
\newblock Sub-graph contrast for scalable self-supervised graph representation
  learning.
\newblock In \emph{IEEE International Conference on Data Mining}, pp.\
  222--231, 2020.

\bibitem[Jin et~al.(2020)Jin, Derr, Liu, Wang, Wang, Liu, and
  Tang]{jin2020self}
Jin, W., Derr, T., Liu, H., Wang, Y., Wang, S., Liu, Z., and Tang, J.
\newblock Self-supervised learning on graphs: Deep insights and new direction.
\newblock \emph{arXiv preprint arXiv:2006.10141}, 2020.

\bibitem[Kim \& Oh(2021)Kim and Oh]{kim2021how}
Kim, D. and Oh, A.
\newblock How to find your friendly neighborhood: Graph attention design with
  self-supervision.
\newblock In \emph{International Conference on Learning Representations}, 2021.

\bibitem[Kingma \& Ba(2014)Kingma and Ba]{kingma2014adam}
Kingma, D.~P. and Ba, J.
\newblock Adam: A method for stochastic optimization.
\newblock \emph{arXiv preprint arXiv:1412.6980}, 2014.

\bibitem[Kipf \& Welling(2016)Kipf and Welling]{kipf2016variational}
Kipf, T.~N. and Welling, M.
\newblock Variational graph auto-encoders.
\newblock \emph{arXiv preprint arXiv:1611.07308}, 2016.

\bibitem[Kipf \& Welling(2017)Kipf and Welling]{kipf2017semi}
Kipf, T.~N. and Welling, M.
\newblock Semi-supervised classification with graph convolutional networks.
\newblock In \emph{International Conference on Learning Representations}, 2017.

\bibitem[Laine et~al.(2019)Laine, Karras, Lehtinen, and Aila]{laine2019high}
Laine, S., Karras, T., Lehtinen, J., and Aila, T.
\newblock High-quality self-supervised deep image denoising.
\newblock \emph{Advances in Neural Information Processing Systems},
  32:\penalty0 6970--6980, 2019.

\bibitem[Liu et~al.(2021{\natexlab{a}})Liu, Luo, Wang, Xie, Yuan, Gui, Yu, Xu,
  Zhang, Liu, Yan, Liu, Fu, Oztekin, Zhang, and Ji]{JMLR:v22:21-0343}
Liu, M., Luo, Y., Wang, L., Xie, Y., Yuan, H., Gui, S., Yu, H., Xu, Z., Zhang,
  J., Liu, Y., Yan, K., Liu, H., Fu, C., Oztekin, B.~M., Zhang, X., and Ji, S.
\newblock {DIG}: A turnkey library for diving into graph deep learning
  research.
\newblock \emph{Journal of Machine Learning Research}, 22\penalty0
  (240):\penalty0 1--9, 2021{\natexlab{a}}.
\newblock URL \url{http://jmlr.org/papers/v22/21-0343.html}.

\bibitem[Liu et~al.(2021{\natexlab{b}})Liu, Zhang, Hou, Mian, Wang, Zhang, and
  Tang]{liu2020self}
Liu, X., Zhang, F., Hou, Z., Mian, L., Wang, Z., Zhang, J., and Tang, J.
\newblock Self-supervised learning: Generative or contrastive.
\newblock \emph{IEEE Transactions on Knowledge and Data Engineering},
  2021{\natexlab{b}}.

\bibitem[McAuley et~al.(2015)McAuley, Targett, Shi, and Van
  Den~Hengel]{mcauley2015image}
McAuley, J., Targett, C., Shi, Q., and Van Den~Hengel, A.
\newblock Image-based recommendations on styles and substitutes.
\newblock In \emph{Proceedings of the 38th International ACM SIGIR Conference
  on Research and Development in Information Retrieval}, pp.\  43--52, 2015.

\bibitem[McCallum et~al.(2000)McCallum, Nigam, Rennie, and
  Seymore]{mccallum2000cora}
McCallum, A.~K., Nigam, K., Rennie, J., and Seymore, K.
\newblock Automating the construction of internet portals with machine
  learning.
\newblock \emph{Information Retrieval}, 3\penalty0 (2):\penalty0 127--163,
  2000.

\bibitem[Mikolov et~al.(2013{\natexlab{a}})Mikolov, Chen, Corrado, and
  Dean]{Mikolov2013EfficientEO}
Mikolov, T., Chen, K., Corrado, G., and Dean, J.
\newblock Efficient estimation of word representations in vector space.
\newblock In \emph{International Conference on Learning Representations},
  2013{\natexlab{a}}.

\bibitem[Mikolov et~al.(2013{\natexlab{b}})Mikolov, Sutskever, Chen, Corrado,
  and Dean]{Mikolov2013DistributedRO}
Mikolov, T., Sutskever, I., Chen, K., Corrado, G., and Dean, J.
\newblock Distributed representations of words and phrases and their
  compositionality.
\newblock In \emph{Advances in Neural Information Processing Systems},
  2013{\natexlab{b}}.

\bibitem[Morris et~al.(2020)Morris, Kriege, Bause, Kersting, Mutzel, and
  Neumann]{Morris2020tudataset}
Morris, C., Kriege, N.~M., Bause, F., Kersting, K., Mutzel, P., and Neumann, M.
\newblock {TUDataset}: A collection of benchmark datasets for learning with
  graphs.
\newblock In \emph{ICML 2020 Workshop on Graph Representation Learning and
  Beyond (GRL+ 2020)}, 2020.

\bibitem[Narayanan et~al.(2017)Narayanan, Chandramohan, Venkatesan, Chen, Liu,
  and Jaiswal]{narayanan2017graph2vec}
Narayanan, A., Chandramohan, M., Venkatesan, R., Chen, L., Liu, Y., and
  Jaiswal, S.
\newblock graph2vec: Learning distributed representations of graphs.
\newblock \emph{arXiv preprint arXiv:1707.05005}, 2017.

\bibitem[Peng et~al.(2020)Peng, Huang, Luo, Zheng, Rong, Xu, and
  Huang]{peng2020graph}
Peng, Z., Huang, W., Luo, M., Zheng, Q., Rong, Y., Xu, T., and Huang, J.
\newblock Graph representation learning via graphical mutual information
  maximization.
\newblock In \emph{Proceedings of The Web Conference 2020}, pp.\  259--270,
  2020.

\bibitem[Perozzi et~al.(2014{\natexlab{a}})Perozzi, Al-Rfou, and
  Skiena]{Perozzi2014deepwalk}
Perozzi, B., Al-Rfou, R., and Skiena, S.
\newblock Deepwalk: Online learning of social representations.
\newblock In \emph{Proceedings of the 20th ACM SIGKDD International Conference
  on Knowledge Discovery and Data Mining}, pp.\  701--710, 2014{\natexlab{a}}.

\bibitem[Perozzi et~al.(2014{\natexlab{b}})Perozzi, Al-Rfou, and
  Skiena]{deepwalk}
Perozzi, B., Al-Rfou, R., and Skiena, S.
\newblock {DeepWalk}: Online learning of social representations.
\newblock In \emph{Proceedings of the 20th ACM SIGKDD International Conference
  on Knowledge Discovery and Data Mining}, pp.\  701--710, 2014{\natexlab{b}}.

\bibitem[Rong et~al.(2020)Rong, Bian, Xu, Xie, Wei, Huang, and
  Huang]{rong2020grover}
Rong, Y., Bian, Y., Xu, T., Xie, W., Wei, Y., Huang, W., and Huang, J.
\newblock Self-supervised graph transformer on large-scale molecular data.
\newblock \emph{Advances in Neural Information Processing Systems}, pp.\
  12559--12571, 2020.

\bibitem[Saxe et~al.(2018)Saxe, Bansal, Dapello, Advani, Kolchinsky, Tracey,
  and Cox]{michael2018on}
Saxe, A.~M., Bansal, Y., Dapello, J., Advani, M., Kolchinsky, A., Tracey,
  B.~D., and Cox, D.~D.
\newblock On the information bottleneck theory of deep learning.
\newblock In \emph{International Conference on Learning Representations}, 2018.

\bibitem[Sen et~al.(2008)Sen, Namata, Bilgic, Getoor, Galligher, and
  Eliassi-Rad]{sen2008pubmed}
Sen, P., Namata, G., Bilgic, M., Getoor, L., Galligher, B., and Eliassi-Rad, T.
\newblock Collective classification in network data.
\newblock \emph{AI magazine}, 29\penalty0 (3):\penalty0 93--93, 2008.

\bibitem[Shchur et~al.(2018)Shchur, Mumme, Bojchevski, and
  G{\"u}nnemann]{shchur2018pitfalls}
Shchur, O., Mumme, M., Bojchevski, A., and G{\"u}nnemann, S.
\newblock Pitfalls of graph neural network evaluation.
\newblock \emph{Relational Representation Learning Workshop, NeurIPS}, 2018.

\bibitem[Shervashidze et~al.(2009)Shervashidze, Vishwanathan, Petri, Mehlhorn,
  and Borgwardt]{shervashidze2009efficient}
Shervashidze, N., Vishwanathan, S., Petri, T., Mehlhorn, K., and Borgwardt, K.
\newblock Efficient graphlet kernels for large graph comparison.
\newblock In \emph{Artificial intelligence and statistics}, pp.\  488--495.
  PMLR, 2009.

\bibitem[Shervashidze et~al.(2011)Shervashidze, Schweitzer, Van~Leeuwen,
  Mehlhorn, and Borgwardt]{shervashidze2011weisfeiler}
Shervashidze, N., Schweitzer, P., Van~Leeuwen, E.~J., Mehlhorn, K., and
  Borgwardt, K.~M.
\newblock {Weisfeiler-Lehman} graph kernels.
\newblock \emph{Journal of Machine Learning Research}, 12\penalty0 (9), 2011.

\bibitem[Sinha et~al.(2015)Sinha, Shen, Song, Ma, Eide, Hsu, and
  Wang]{sinha2015overview}
Sinha, A., Shen, Z., Song, Y., Ma, H., Eide, D., Hsu, B.-J., and Wang, K.
\newblock An overview of microsoft academic service (mas) and applications.
\newblock In \emph{Proceedings of the 24th International Conference on World
  Wide Web}, pp.\  243--246, 2015.

\bibitem[Sun et~al.(2019)Sun, Hoffman, Verma, and Tang]{sun2019infograph}
Sun, F.-Y., Hoffman, J., Verma, V., and Tang, J.
\newblock Infograph: Unsupervised and semi-supervised graph-level
  representation learning via mutual information maximization.
\newblock In \emph{International Conference on Learning Representations}, 2019.

\bibitem[Thakoor et~al.(2021)Thakoor, Tallec, Azar, Munos,
  Veli{\v{c}}kovi{\'c}, and Valko]{thakoor2021bootstrapped}
Thakoor, S., Tallec, C., Azar, M.~G., Munos, R., Veli{\v{c}}kovi{\'c}, P., and
  Valko, M.
\newblock Bootstrapped representation learning on graphs.
\newblock In \emph{ICLR 2021 Workshop on Geometrical and Topological
  Representation Learning}, 2021.

\bibitem[Tishby \& Zaslavsky(2015)Tishby and Zaslavsky]{tishby2015deep}
Tishby, N. and Zaslavsky, N.
\newblock Deep learning and the information bottleneck principle.
\newblock In \emph{2015 IEEE Information Theory Workshop (ITW)}, pp.\  1--5.
  IEEE, 2015.

\bibitem[Tishby et~al.(1999)Tishby, Pereira, and Bialek]{tishby2000information}
Tishby, N., Pereira, F.~C., and Bialek, W.
\newblock The information bottleneck method.
\newblock In \emph{The 37th annual Allerton Conference on Communication,
  Control, and Computing}, pp.\  368--377, 1999.

\bibitem[Ulyanov et~al.(2018)Ulyanov, Vedaldi, and Lempitsky]{ulyanov2018deep}
Ulyanov, D., Vedaldi, A., and Lempitsky, V.
\newblock Deep image prior.
\newblock In \emph{Proceedings of the IEEE conference on computer vision and
  pattern recognition}, pp.\  9446--9454, 2018.

\bibitem[Veli{\v c}kovi{\'c} et~al.(2019)Veli{\v c}kovi{\'c}, Fedus, Hamilton,
  Li{\`o}, Bengio, and Hjelm]{velikovi2019deep}
Veli{\v c}kovi{\'c}, P., Fedus, W., Hamilton, W.~L., Li{\`o}, P., Bengio, Y.,
  and Hjelm, D.
\newblock Deep graph infomax.
\newblock In \emph{International Conference on Learning Representations}, 2019.

\bibitem[Vincent et~al.(2010)Vincent, Larochelle, Lajoie, Bengio, Manzagol, and
  Bottou]{vincent2010stacked}
Vincent, P., Larochelle, H., Lajoie, I., Bengio, Y., Manzagol, P.-A., and
  Bottou, L.
\newblock Stacked denoising autoencoders: Learning useful representations in a
  deep network with a local denoising criterion.
\newblock \emph{Journal of Machine Learning Research}, 11\penalty0 (12), 2010.

\bibitem[{Wale} \& {Karypis}(2006){Wale} and {Karypis}]{NCI1}
{Wale}, N. and {Karypis}, G.
\newblock Comparison of descriptor spaces for chemical compound retrieval and
  classification.
\newblock In \emph{Sixth International Conference on Data Mining}, pp.\
  678--689, 2006.

\bibitem[Wang et~al.(2017)Wang, Pan, Long, Zhu, and Jiang]{wang2017mgae}
Wang, C., Pan, S., Long, G., Zhu, X., and Jiang, J.
\newblock {MGAE}: Marginalized graph autoencoder for graph clustering.
\newblock In \emph{Proceedings of the 2017 ACM on Conference on Information and
  Knowledge Management}, pp.\  889--898, 2017.

\bibitem[Wang et~al.(2019)Wang, Wang, Xiong, Yu, Guo, Chang, and
  Wang]{wang2019self}
Wang, H., Wang, X., Xiong, W., Yu, M., Guo, X., Chang, S., and Wang, W.~Y.
\newblock Self-supervised learning for contextualized extractive summarization.
\newblock In \emph{Proceedings of the 57th Annual Meeting of the Association
  for Computational Linguistics}, pp.\  2221--2227, 2019.

\bibitem[Wei et~al.(2021)Wei, Shen, Chen, and Ma]{wei2021theoretical}
Wei, C., Shen, K., Chen, Y., and Ma, T.
\newblock Theoretical analysis of self-training with deep networks on unlabeled
  data.
\newblock In \emph{International Conference on Learning Representations}, 2021.

\bibitem[Wu et~al.(2019)Wu, Wang, and Wang]{wu2019self}
Wu, J., Wang, X., and Wang, W.~Y.
\newblock Self-supervised dialogue learning.
\newblock In \emph{Proceedings of the 57th Annual Meeting of the Association
  for Computational Linguistics}, pp.\  3857--3867, 2019.

\bibitem[Xie et~al.(2013)Xie, Kelley, and Szymanski]{xie2013overlapping}
Xie, J., Kelley, S., and Szymanski, B.~K.
\newblock Overlapping community detection in networks: The state-of-the-art and
  comparative study.
\newblock \emph{ACM Computing Surveys (CSUR)}, 45\penalty0 (4):\penalty0 1--35,
  2013.

\bibitem[Xie et~al.(2020)Xie, Wang, and Ji]{XieNoise2Same}
Xie, Y., Wang, Z., and Ji, S.
\newblock {Noise2Same}: Optimizing a self-supervised bound for image denoising.
\newblock In \emph{Proceedings of the 34th Annual Conference on Neural
  Information Processing Systems}, pp.\  20320--20330, 2020.

\bibitem[Xie et~al.(2022)Xie, Xu, Zhang, Wang, and Ji]{xie2022self}
Xie, Y., Xu, Z., Zhang, J., Wang, Z., and Ji, S.
\newblock Self-supervised learning of graph neural networks: A unified review.
\newblock \emph{IEEE Transactions on Pattern Analysis and Machine
  Intelligence}, 2022.

\bibitem[Xu et~al.(2019)Xu, Hu, Leskovec, and Jegelka]{xu2018how}
Xu, K., Hu, W., Leskovec, J., and Jegelka, S.
\newblock How powerful are graph neural networks?
\newblock In \emph{International Conference on Learning Representations}, 2019.

\bibitem[Yanardag \& Vishwanathan(2015)Yanardag and
  Vishwanathan]{yanardag2015deep}
Yanardag, P. and Vishwanathan, S.
\newblock Deep graph kernels.
\newblock In \emph{Proceedings of the 21th ACM SIGKDD International Conference
  on Knowledge Discovery and Data Mining}, pp.\  1365--1374, 2015.

\bibitem[You et~al.(2020)You, Chen, Sui, Chen, Wang, and Shen]{you2020graph}
You, Y., Chen, T., Sui, Y., Chen, T., Wang, Z., and Shen, Y.
\newblock Graph contrastive learning with augmentations.
\newblock In \emph{Advances in Neural Information Processing Systems},
  volume~33, pp.\  5812--5823, 2020.

\bibitem[Zbontar et~al.(2021)Zbontar, Jing, Misra, LeCun, and
  Deny]{zbontar2021barlow}
Zbontar, J., Jing, L., Misra, I., LeCun, Y., and Deny, S.
\newblock Barlow twins: Self-supervised learning via redundancy reduction.
\newblock In \emph{International Conference on Machine Learning}, pp.\
  12310--12320. PMLR, 2021.

\bibitem[Zeng et~al.(2020)Zeng, Zhou, Srivastava, Kannan, and
  Prasanna]{zeng2019graphsaint}
Zeng, H., Zhou, H., Srivastava, A., Kannan, R., and Prasanna, V.
\newblock {GraphSAINT}: Graph sampling based inductive learning method.
\newblock In \emph{International Conference on Learning Representations}, 2020.

\bibitem[Zhu et~al.(2020)Zhu, Xu, Yu, Liu, Wu, and Wang]{zhu2020deep}
Zhu, Y., Xu, Y., Yu, F., Liu, Q., Wu, S., and Wang, L.
\newblock Deep graph contrastive representation learning.
\newblock In \emph{ICML Workshop on Graph Representation Learning and Beyond},
  2020.

\bibitem[Zhu et~al.(2021)Zhu, Xu, Yu, Liu, Wu, and Wang]{zhu2021graph}
Zhu, Y., Xu, Y., Yu, F., Liu, Q., Wu, S., and Wang, L.
\newblock Graph contrastive learning with adaptive augmentation.
\newblock In \emph{Proceedings of the Web Conference 2021}, pp.\  2069--2080,
  2021.

\bibitem[Zitnik \& Leskovec(2017)Zitnik and Leskovec]{zitnik2017predicting}
Zitnik, M. and Leskovec, J.
\newblock Predicting multicellular function through multi-layer tissue
  networks.
\newblock \emph{Bioinformatics}, 33\penalty0 (14):\penalty0 i190--i198, 2017.

\end{thebibliography}

\newpage
\appendix
\onecolumn
\section{Proof of Theorem 1}\label{sec:proof1}
\begin{proof}
We first derive the relationship between the supervised objective of latent graph prediction $\mathds{E}\norm{f(\bm A, \bm X)-\bm F}^2$ and the self-supervised reconstruction loss $\mathds{E}\norm{f(\bm{A},\bm X)-\bm{X}}^2$ in the following equations,
\begin{align}
    \mathds{E}\norm{f(\bm{A},\bm X)-\bm{X}}^2 &= \mathds{E}\norm{(f(\bm{A},\bm X)-\bm F) - (\bm X -\bm F)}^2\\
    = \mathds{E}\big[&\norm{f(\bm{A},\bm X)-\bm F}^2+\norm{\bm X-\bm F}^2-2\inner{f(\bm{A},\bm X)-\bm F, \bm X - \bm F}\big],
\end{align}
where $\inner{\cdot, \cdot}$ denotes the inner product along all dimensions $\{0, \cdots, d|V|-1\}$. The expectation $\mathds{E}{\norm{f(\bm{A},\bm X)}}$ in the above equation is not relevant to the neural network $f$. It hence can be considered as a constant during the optimization of $f$. To derive an upper bound to $\mathds{E}\norm{f(\bm A, \bm X)-\bm F}^2$, we only need to derive an upper bound of its equivalent $\mathds{E}\norm{f(\bm{A},\bm X)-\bm{X}}^2+2\mathds{E}\inner{f(\bm{A},\bm X)-\bm F, \bm X-\bm F}$. As $\bm F$ is unobserved, our goal is to derive an upper bound to eliminate the need of $\bm F$ for the inner product term $\inner{f(\bm{A},\bm X)-\bm F, \bm X-\bm F}$. To do so, we apply the definition of latent graph $\mathds{E}[\bm X|\bm A, \bm F]=\bm F$ and rewrite the inner product into the following form.
\begin{align}
    \mathds{E}\inner{f(\bm{A},\bm X)-\bm F, \bm X-\bm F}
    &= \mathds{E}_{\bm A, \bm F}\mathds{E}_{\bm X}\bigg[\sum_i(f_i(\bm A, \bm{X})-{\bm F}_i)(\bm X_i-\bm F_i)|\bm A, \bm F\bigg]\\
    &= \sum_i\mathds{E}_{\bm A, \bm F}\bigg[\mathds{E}\big[(f_i(\bm A, \bm{X})-\bm F_i)(\bm X_i-\bm F_i)|\bm A, \bm F\big]-\nonumber\\
    &\quad\quad\quad\quad\quad\quad\quad\quad\mathds{E}\big[f_i(\bm A, \bm{X})-\bm F_i|\bm F\big]\mathds{E}\big[\bm X_i-\bm F_i|\bm A, \bm F\big]\bigg]\\
    &= \sum_i\mathds{E}_{\bm A, \bm F}\left[\mathrm{Cov}(f_i(\bm A, \bm X)-\bm F_i, \bm X_i-\bm F_i|\bm A, \bm F) \right]\\
    &= \sum_i\mathds{E}_{\bm A, \bm F}\left[\mathrm{Cov}(f_i(\bm A, \bm X), \bm X_i|\bm A, \bm F) \right],
\end{align}
where $i$ sums over all dimensions $\{0, \cdots, d|V|-1\}$, $f_i$ and $\bm X_i$ denotes the $i$-th element of the flattened matrices. Note that we employ $\mathds{E}[\bm X-\bm F|\bm A, \bm F]=\bm0$ to let Equation~(17) hold, according to the definition of latent graphs. Letting $J$ be a uniformly sampled subset of all node indices $\{0, \cdots, |V|-1\}$, the right hand side of the above equation satisfies
\begin{equation}
    \mbox{RHS} = \mathds{E}_J\frac{|V|}{|J|}\sum_{j\in J}\sum_{k=0}^{d-1}\mathds{E}_{\bm A, \bm F}\left[\mathrm{Cov}(f_{jd+k}(\bm A, \bm X), \bm X_{jd+k}|\bm A,\bm F) \right],
\end{equation}
where $f_{jd+k}\in\mathds{R}$ and $X_{jd+k}\in\mathds{R}$ denote the $(jd+k)$-th element of corresponding matrices, \textit{i.e.}, the $k$-th element of the node $v_j$, whereas $X_J\in\mathds{R}^{|J|\times d}$ denotes the feature matrix of nodes in $V_J$. Given the bounded variance $\mathrm{Var}(\bm X_i)\le\sigma^2,\forall i$, we bound the above term as
\begin{align}
    \mbox{RHS}
    &= \mathds{E}_J\frac{|V|}{|J|}\sum_{j\in J, k}\mathds{E}_{\bm A, \bm F}\Big[\mathrm{Cov}(f_{jd+k}(\bm A, \bm X)-f_{jd+k}(\bm A, \bm X_{J^c}), \bm X_{jd+k}|\bm A,\bm F)\Big] \\
    &\le \mathds{E}_J\frac{|V|}{|J|}\sum_{j\in J, k}\mathds{E}_{\bm A, \bm F}\Big[\mathrm{Var}(f_{jd+k}(\bm A, \bm X)-f_{jd+k}(\bm A, \bm X_{J^c})|\bm A,\bm F)\cdot \mathrm{Var}(\bm X_{jd+k})\Big]^{1/2} \\
    &\le |V|\mathds{E}_J\Bigg(\frac{1}{|J|}\sum_{j\in J, k}\mathds{E}_{\bm A, \bm F}\Big[\mathrm{Var}(f_{jd+k}(\bm A, \bm X)-f_{jd+k}(\bm A, \bm X_{J^c})|\bm A,\bm F)\cdot\sigma^2\Big]\Bigg)^{1/2} \\
    &\le \sigma|V|\mathds{E}_J\Bigg(\frac{1}{|J|}\sum_{j\in J, k}\mathds{E}_{\bm A, \bm F}\Bigg[\mathds{E}\Big[\big[f_{jd+k}(\bm A, \bm X)-f_{jd+k}(\bm A, \bm X_{J^c})\big]^2|\bm A,\bm F\Big]\Bigg]\Bigg)^{1/2}\\
    &= \sigma|V|\mathds{E}_J\Bigg(\frac{1}{|J|}\sum_{j\in J, k}\mathds{E}\Big[f_{jd+k}(\bm A, \bm X)-f_{jd+k}(\bm A, \bm X_{J^c})\Big]^2\Bigg)^{1/2}\\
    &= \sigma|V|\mathds{E}_J\left(\frac{1}{|J|}\mathds{E}\norm{f_J(\bm A, \bm X)-f_J(\bm A, \bm X_{J^c})}^2\right)^{1/2}.
\end{align}
Above inequalities and equations are derived based on the fact that $f_J(\bm A, \bm X_{J^c})$ does not correlate to $\bm X_{jd+k}$ as $j\notin J^c$ for Equation~(21), the Cauchy-Schwarz inequality for Inequality~(22), and $(\mathds{E}X)^2\le\mathds{E}X^2$ for Inequality~(23). We complete the proof of Theorem 1 by combining Equation~(15) and Inequality~(26),
\begin{align}
    \mathds{E}\big[\norm{f(\bm{A},\bm X)-\bm F}^2+\norm{\bm X-\bm F}^2\big] &= \mathds{E}\norm{f(\bm{A},\bm X)-\bm{X}}^2 + 2\inner{f(\bm{A},\bm X)-\bm F, \bm X - \bm F}\\
    \le \mathds{E}\norm{f(\bm{A},\bm X)-\bm{X}}^2 + & 2\sigma|V|\mathds{E}_J\left(\frac{1}{|J|}\mathds{E}\norm{f_J(\bm A, \bm X)-f_J(\bm A, \bm X_{J^c})}^2\right)^{1/2}.
\end{align}
\end{proof}

\section{Proof of Corollary 1 and 2}\label{sec:proof2}
\begin{proof}
We first prove Corollary 1. Consider an $\ell$-Lipschitz continuous prediction head with respect to $l_2$-norm consists of fully connected layers. We have
\begin{align}
    \norm{f_J(\bm A, \bm X)-f_J(\bm A, \bm X_{J^c})}_2 &= \norm{\mathcal{D}(\bm H_J)-\mathcal{D}(\bm H'_J)}_2
    \le \ell\norm{\bm H_J-\bm H'_J}_2.
\end{align}
We therefore have the following inequality
\begin{align}
    \mathds{E}\norm{f_J(\bm A, \bm X)-f_J(\bm A, \bm X_{J^c})}_2^2 &
    \le \mathds{E}\bigg[\ell^2\norm{\bm H_J-\bm H'_J}_2^2\bigg].
\end{align}
We apply the above inequality to Theorem~1 and obtain the following inequality
\begin{align}
    \mathds{E}\big[\norm{f(\bm{A},\bm X)-\bm F}^2+\norm{\bm X-\bm F}^2\big]&\nonumber\\
    \le \mathds{E}\norm{f(\bm{A},\bm X)-\bm{X}}^2 +&  2\sigma|V|\mathds{E}_J\bigg(\frac{1}{|J|}\mathds{E}\norm{f_J(\bm A, \bm X)-f_J(\bm A, \bm X_{J^c})}^2\bigg)^{1/2}\\
    \le \mathds{E}\norm{f(\bm{A},\bm X)-\bm{X}}^2 +&  2\sigma|V|\mathds{E}_J\bigg(\frac{1}{|J|}\mathds{E}\bigg[\ell^2\norm{\bm H_J-\bm H'_J}_2^2\bigg]\bigg)^{1/2}\\
    = \mathds{E}\norm{f(\bm{A},\bm X)-\bm{X}}^2 +&  2\sigma|V|\ell\mathds{E}_J\bigg(\mathds{E}\norm{\bm H_J-\bm H'_J}_2^2/|J|\bigg)^{1/2},
\end{align}
which completes the proof of Corollay~1.

Similarly, for Corollay~2, we have
\begin{align}
    \norm{f_J(\bm A, \bm X)-f_J(\bm A, \bm X_{J^c})}_2 &= \norm{\mathcal{D}(\bm H_J)-\mathcal{D}(\bm H'_J)}_2
    \le \ell\norm{\bm H_J-\bm H'_J}_2.
\end{align}
Given an $\ell_r$-Bilipschitz continuous readout function $\mathcal{R}$, the following inequalities hold,
\begin{align}
    \frac{1}{\ell_r}\norm{\bm H_J-\bm H'_J}_2
    \le\norm{\mathcal{R}(\bm H_J)-\mathcal{R}(\bm H'_J)}_2
    \le \ell_r\norm{\bm H_J-\bm H'_J}_2.
\end{align}
We therefore have 
\begin{align}
    \mathds{E}\big[\norm{f(\bm{A},\bm X)-\bm F}^2+\norm{\bm X-\bm F}^2\big]&\nonumber\\
    \le \mathds{E}\norm{f(\bm{A},\bm X)-\bm{X}}^2 +&  2\sigma|V|\ell\mathds{E}_J\bigg(\mathds{E}\norm{\bm H_J-\bm H'_J}_2^2/|J|\bigg)^{1/2}\\
    \le \mathds{E}\norm{f(\bm{A},\bm X)-\bm{X}}^2 +&  2\sigma|V|\ell\ell_r\mathds{E}_J\bigg(\mathds{E}\norm{\mathcal{R}(\bm H_J)-\mathcal{R}(\bm H'_J)}_2^2/|J|\bigg)^{1/2}\\
    = \mathds{E}\norm{f(\bm{A},\bm X)-\bm{X}}^2 +&  2\sigma|V|k\ell\mathds{E}_J\bigg(\mathds{E}\norm{\bm z - \bm z'}_2^2/|J|\bigg)^{1/2},
\end{align}
which completes the proof of Corollay~2.
\end{proof}

\section{Experiment Settings and Model Configurations}\label{sec:config}

\red{\paragraph{Dataset Statistics.} Statistics including number of graphs, averaged number of nodes, averaged number of edges, and node attribute dimensions are summarized in Table~4.}

\begin{table*}[th]\label{tab:sum}
\caption{Summary and statistics of common graph datasets for self-supervised learning.}
\centering
\begin{tabular}{lcccccc}
\toprule
Datasets & Evaluation task & \# graphs & Avg. nodes & Avg. edges & \# features     \\ \hline
\textbf{NCI1}     & \multirow{8}{*}{\begin{tabular}[c]{@{}c@{}}Graph-level \\classification\end{tabular}}  & 4110      & 29.87      & 32.30      & 37              \\
\textbf{PROTEINS} &                                                                                      & 1178      & 39.06      & 72.82      & 3              \\
\textbf{DD}       &                                                                                      & 188       & 284.32     & 715.66     & 89              \\
\textbf{MUTAG}    &                                                                                      & 1113      & 17.93      & 19.79      & 7             \\
\textbf{COLLAB}   &                                                                                      & 5000      & 74.49      & 2457.78    & 1              \\
\textbf{RDT-B}    &                                                                                      & 2000      & 429.63     & 497.75     & 1              \\
\textbf{RDT-M5K}  &                                                                                      & 4999      & 508.52     & 594.87     & 1              \\
\textbf{IMDB-B}   &                                                                                      & 1000      & 19.77      & 96.53      & 1              \\ \hline
\textbf{Amazon Computer}    & \multirow{4}{*}{\begin{tabular}[c]{@{}c@{}}Transductive \\Node-level \\classification\end{tabular}}   & 1         & 13,752     & 245,861      & 767              \\
\textbf{Amazon Photos}      &                                                                                                   & 1         & 7,650      & 119,081      & 745              \\
\textbf{Coauthor CS}        &                                                                                                   & 1         & 81,894     & 81,894       & 6,806            \\ 
\textbf{Coauthor Physics}   &                                                                                                   & 1         & 247,962    & 247,962      & 8,415            \\ \hline
\textbf{PPI}                & \multirow{3}{*}{\begin{tabular}[c]{@{}c@{}}Inductive \\Node-level \\classification\end{tabular}}      & 24        & 2,373      & 34,133       & 50               \\
\textbf{Flickr}             &                                                                                                   & 1         & 89,250     & 899,756      & 500               \\
\textbf{Reddit}             &                                                                                                   & 1         & 232,965    & 11,606,919   & 602               \\ \bottomrule
\end{tabular}
\end{table*}

\paragraph{Models for Graph-Level Datasets.} We employ a 3-layer GIN~\citep{xu2018how} as the graph encoder $\mathcal{E}$, and a 2-layer MLP as the decoder $\mathcal{D}$. Following GraphCL~\citep{you2020graph}, we use a hidden dimension of size 32 and concatenate the embedding at each encoding layer to obtain the final representation. To fulfill the conditions in Corollary~2, we apply global sum pooling as the readout function $\mathcal{R}$. The obtained graph representation is then taken by a SVM classifier with a 10-fold evaluation. For graph datasets that do not come with node attributes, we apply the one-hot vector of the degree for each node as the node attributes so that the node degrees are reconstructed. Certain thresholds for max degrees are applied to reduce computational cost and avoid over sparse node features. The neural network is trained using the loss described in Equation~(6).
We mask all attributes of the sampled nodes with Gaussian noise.
Detailed training configurations including mask ratio, the standard deviation of noise, weight scalar $\alpha'$, and threshold for max degrees are shown in Table~\ref{tab:config_graph}. \red{Note that we do not include carefully designed implementation mechanisms by BGRL, such as stop gradients, EMA, and batch normalization at the last layer.}

\paragraph{Models for Node-Level Datasets.} For node-level datasets, we employ a 2-layer GCN~\citep{kipf2017semi} as the graph encoder $\mathcal{E}$, and a linear layer or an MLP as the decoder $\mathcal{D}$. We use a hidden dimension of size 512 at each encoding layer. The neural network is trained using the loss described in Equation~(3).
We uniformly employ the weight scalar $\alpha'$ of $2$ as we observed that the model performance is not sensitive to the selection of $\alpha'$ within the range $[1,100]$. We obtain the final node representation by concatenating the original feature with the embedding from the last layer of the encoder.
\red{The intuition of this is based on the Bayesian rule where the learned encoder provides the prior knowledge~\citep{ulyanov2018deep} of data distribution whereas the given graph data serves as the observed samples. And the posteriori should be based on a combination of the priori (encoder output) and the observed data itself~\citep{laine2019high}.}
Node representation is then taken by a logistic regression classifier that is trained using the cross-entropy (CE) loss with a learning rate of $0.01$. 
Detailed training configurations including mask ratio, the standard deviation of noise, number of encoder and decoder layers, learning rate and weight decay of the graph neural network, training epochs, and weight decay of the logistic regression classifier are shown in Table~\ref{tab:config_node}. To split train, valid and test sets, we use the public split used in ~\citep{shchur2018pitfalls} for Coauthor and Amazon, ~\citep{zitnik2017predicting, hamilton2017inductive, zeng2019graphsaint} for PPI, Reddit and Flickr provided by PyTorch Geometric. \red{Note that we do not include implementation mechanisms by BGRL, such as stop gradients, EMA, and batch normalization at the last layer.}

\begin{table}[bh]
    \small
    \centering
    \addtolength{\tabcolsep}{-2pt}
    \caption{Model configurations for graph-level datasets.}
    \begin{tabular}{ccccccccc}
    \toprule
                     & NCI1 & PROTEINS & DD  & MUTAG & COLLAB & RDT-B & RDT-M5K & IMDB-B \\ 
    \hline
    Mask ratio       & 0.05 & 0.3      & 0.1 & 0.05  & 0.05   & 0.05  & 0.05    & 0.05   \\
    Noise SD         & 0.5  & 2        & 0.5 & 0.5   & 0.5    & 0.5   & 0.5     & 0.5    \\
    Weight scalar $\alpha'$    & 10  & 1        & 10  & 10   & 10     & 10    & 10      & 10     \\
    Degree threshold & --    & --        & --   & --     & 128    & --     & --       & 64     \\
    Learning rate    & $10^{-5}$  & $10^{-5}$  & $10^{-5}$  & $10^{-5}$  & $10^{-4}$  & $10^{-3}$  & $10^{-4}$  & $10^{-4}$ \\
    \bottomrule
    \end{tabular}
    \label{tab:config_graph}
\end{table}

\begin{table}[H]
    \small
    \centering
    \addtolength{\tabcolsep}{-2pt}
    \caption{Model configurartions for node-level datasets.}
    \begin{tabular}{cccccccc}
    \toprule
                          & Am.Computers & Am.Photos & CoautherCS & CoauthorPhy & PPI      & Flickr   & Reddit   \\
    \hline
    Mask ratio             & 0.05          & 0.05      & 0.05      & 0.05        & 0.05     & 0.01     & 0.05     \\
    Noise SD               & 0.5          & 0.5       & 0.005      & 0.5         & 0.5      & 0.5      & 0.5      \\
    Decoder layers       & 1            & 1         & 1          & 2           & 2        & 2        & 2        \\
    Learning rate          & $10^{-4}$     & $10^{-5}$  & $10^{-3}$   & $10^{-3}$    & $10^{-3}$ & $10^{-4}$ & $10^{-3}$ \\
    Weight decay & 0            & $10^{-4}$  & 0          & 0           & $10^{-5}$ & 0        & 0        \\
    LogReg epochs          & 400          & 400       & 400        & 300         & 200      & 200      & 500      \\
    LogReg WD    & $10^{-3}$     & $10^{-3}$  & $10^{-3}$   & $10^{-3}$    & 0        & $10^{-3}$ & $10^{-3}$ \\
    \bottomrule
    \end{tabular}
    \label{tab:config_node}
\end{table}

\section{Experimental Results under Semi-supervised Setting}\label{sec:semi}
For graph-level datasets, we perform semi-supervised experiments with 10\% label rate using both GIN and GCN. All experiments are conducted with the same random seed to avoid randomness in data split and initialization. Under the setting of random initialization followed by supervised learning, the GNN is randomly initialized without pre-training. Under the setting of LaGraph followed by supervised learning, the GNN is pre-trained with the proposed LaGraph framework. Weights of GNNs are fine-tuned during supervised learning with 10\% labels. For each dataset, the learning rate and epoch number for pre-training are the same as what we use under an unsupervised setting. For fine-tuning, learning rate is selected from $\{10^{-3}, 10^{-4}, 10^{-5}\}$, and epoch number is selected from $\{5, 10, 15, 20\}$. The results shown in Table~\ref{tab:semi_gin} and Table~\ref{tab:semi_gcn} indicate that our proposed LaGraph framework is also effective for semi-supervised learning with different GNN backbones. 

\begin{table}[H]
    \small
    \centering
    \addtolength{\tabcolsep}{-4pt}
    \caption{GIN results for Semi-supervised learning.}
    \begin{tabular}{lcccccccc}
    \toprule
                     & NCI1 & PROTEINS & DD  & MUTAG & COLLAB & RDT-B & RDT-M5K & IMDB-B \\ 
    \hline
    Rand. Init. + 10\% supervised       & 76.67 & 75.29 & 76.66 & 86.67 & 77.54 & 85.45 & 56.03 & 72.70   \\
    LaGraph + 10\% supervised           & 80.19 & 76.10 & 77.93 & 91.40 & 78.04 & 89.65 & 56.43 & 74.30    \\
    \bottomrule
    \end{tabular}
    \label{tab:semi_gin}
\end{table}

\begin{table}[H]
    \small
    \centering
    \addtolength{\tabcolsep}{-4pt}
    \caption{GCN results for Semi-supervised learning.}
    \begin{tabular}{lcccccccc}
    \toprule
                     & NCI1 & PROTEINS & DD  & MUTAG & COLLAB & RDT-B & RDT-M5K & IMDB-B \\ 
    \hline
    Rand. Init. + 10\% supervised       & 75.47 & 74.48 & 77.33 & 84.59 & 79.02 & 85.30 & 53.67 & 72.90   \\
    LaGraph + 10\% supervised           & 78.18 & 76.28 & 78.86 & 85.12 & 80.12 & 90.35 & 55.33 & 75.10    \\
    \bottomrule
    \end{tabular}
    \label{tab:semi_gcn}
\end{table}

\section{Additional Ablation Studies.}\label{sec:ablat}
\textbf{Ablation on Optimizing Different Objectives.} We empirically compare the effect of different upper bounds on graph-level datasets. In addition to the objectives described in Corollary~2, we further train the graph encoder with the upper bound described in Theorem~1, which applies invariance regularization on the reconstructed node features. In addition, as node attributes in many graph-level datasets are formed as one-hot vectors of the node type, we also provide the results of using two corresponding multinomial versions of the objective. In particular, we replace the reconstruction term by the cross-entropy between $f(\bm A, \bm X)$ and $\bm X$ and, if computed on the outputs, the invariance term by the KL-divergence between $f_J(\bm A, \bm X)$ and $f_J(\bm A, \bm X_{J^c})$. Note that the multinomial versions are no longer strictly upper bounds of supervised latent graph prediction. In Table~\ref{tab:objs}, we show the results obtained under the four objectives above, namely, to compute invariance on on-embedding (MSE-Embed), on-output (MSE-Output), and their corresponding multinomial versions (CE-Embed and CE-Output), respectively. Results indicate that there is no significant difference among the four versions on most datasets, while MSE-Embed and CE-Embed generally tend to be more stable and achieve higher performance on MUTAG, RDT-B, and RDT-M5K.

\begin{table}[H]
    \small
    \centering
    \caption{Effect of training with different objectives on graph-level datasets.}\label{tab:objs}
    \addtolength{\tabcolsep}{-3pt}
    \begin{tabular}{ccccccccc}
    \toprule
                  & NCI1       & PROTEINS   & DD         & MUTAG      & COLLAB     & RDT-B      & RDT-M5K    & IMDB-B     \\
    \hline
    MSE-Embed & \textbf{79.9±0.5} & 75.2±0.3 & \textbf{78.1±0.3} & \textbf{90.2±1.3} & 77.6±0.1 & 90.4±0.9 & \textbf{56.4±0.2} & \textbf{73.7±0.7} \\
    MSE-Output & 79.9±0.7 & 75.0±0.4 & 78.1±0.8 & 89.2±2.1 & \textbf{77.7±0.1} & 89.8±0.9 & 56.0±0.4 & 73.4±0.6 \\
    CE-Embed & 79.9±0.5 & \textbf{75.2±0.3} & 78.1±0.4 & 90.1±1.0 & 77.6±0.2 & \textbf{90.5±1.3} & 56.3±0.4 & 73.7±0.7 \\
    CE-Output & 79.9±0.7 & 75.2±0.4 & 78.1±0.4 & 89.3±2.7 & 77.6±0.1 & 89.4±1.8 & 55.7±0.2 & 73.5±0.5 \\
    \bottomrule
    \end{tabular}
    \label{tab:objective}
\end{table}

\red{\textbf{Ablation on Concatenated Representations for Node-Level Datasets.} For the node-level datasets, We obtain the final node representation by concatenating the original feature with the embedding from the last layer of the encoder, due to the intuition discussed in Appendix~\ref{sec:config}. Results in Table~\ref{tab:concat} compare the performance of representations with or without concatenations. The removal of the concatenation leads to reduced performance on four of the seven datasets and performance gain on the rest datasets including the most challenging PPI. The results indicate that the concatenation generally positively contributes to the final performance. However, the conclusion still holds that, on node-level datasets, LaGraph can provide significant performance gain on challenging datasets where there is a gap between SSL and supervised performance. Meanwhile, the performance of LaGraph is on par with the performance of supervised learning and the SOTA method BGRL on datasets that are less challenging.}

\begin{table}[H]
    \small
    \centering
    \caption{Effect of performing concatenation with node features.}
    \begin{tabular}{cccccccc}
    \toprule
    Dataset & Am. Comp. & Am. Pht. & Co. CS & Co. Phy & PPI & Flickr & Reddit \\
    \hline
    With concat & 88.0±0.3 & \textbf{93.5±0.4} & \textbf{93.3±0.2} & \textbf{95.8±0.1} & 74.6±0.0 & 51.3±0.1 & \textbf{95.2±0.0} \\
    W/o concat & \textbf{88.8±0.3} & 92.7±0.4 & 92.6±0.2 & 95.3±0.1 & \textbf{75.2±0.0} & \textbf{51.6±0.1} & 94.8±0.0 \\
    \bottomrule
    \end{tabular}
    \label{tab:concat}
\end{table}

\section{Additional Discussion on Potential Limitations and Future directions}\label{sec:fut}
\red{In this section, we discuss several limitations of the proposed method and their solutions or related future directions.}

\red{\paragraph{Performance comparison with BGRL on transductive tasks.} From the experimental perspective, we admit that BGRL is a quite strong baseline method for transductive tasks. As the results are already on the same level as the performance of supervised training, it is very difficult to further obtain significant improvements. However, when it comes to inductive tasks, where there is still a significant gap between the performance of BGRL and supervised learning, our method is able to bring significant improvements in performance. Therefore, we argue that the non-significant performance boost on some transductive datasets does not degrade our main conclusion about the effectiveness of our method and the contribution of our work.}

\red{\paragraph{Unattributed graphs.} Although, in this work, we mainly focus on graphs with attributed nodes, there exist cases where the nodes are unattributed and all information is contained in the graph topology, especially for some graph-level datasets. In such cases, we follow a common solution to consider the one-hot vectors of node degrees as the node attributes and our objective performs reconstruction on the node degree. To avoid inconsistency between training and testing graphs in their range of degrees, we introduce thresholds to the node degrees, \textit{i.e.}, the degree of a node is considered as $k$ if it exceeds $k$. The current solution can capture the topological information of a graph to some degree. However, there can be better solutions capturing full topological information of graphs. A potential direction is to perform connectivity reconstruction with the invariance of representations to the changing in the input edge set. Although the described approach does not currently fit into our theoretical framework, it is possible to derive similar objectives (e.g., upper bound to link prediction objective) following a similar idea.}

\red{\paragraph{Scaling-up issue.} The scaling-up of graph neural networks becomes an emerging topic. Many existing self-supervised methods may suffer from the scaling-up issue when the graph scales up to billions of nodes and edges. Although we do not perform experimental studies on extremely large graphs, we perform ablation studies to demonstrate the robustness of our method to the training schemes of sampling subgraphs (mini-batches of nodes) for each training iteration.}

\red{\paragraph{Performance of SSL methods on unsupervised downstream tasks.} Performing linear evaluation with supervised downstream tasks on learned representations is the most common way to evaluate the performance of SSL methods. However, evaluation performance on graph-specific unsupervised tasks such as overlapping community detection~\citep{xie2013overlapping} is seldom studied. Further investigations in the unsupervised downstream task are required to fully demonstrate the effectiveness of self-supervised learning methods.}

\end{document}